# Multi-agent self-triage system with medical flowcharts

Yujia Liu[1,2], Sophia Yu[3], Hongyue Jin[1], Jessica Wen[4], Alexander Qian[5], Terrence Lee[6], Mattheus Ramsis[7], Gi Won Choi[8], Lianhui Qin[3], Xin Liu[9,10], Edward J. Wang[1,2]

[1] Department of Electrical and Computer Engineering, University of California San Diego, La Jolla, CA, USA.
[2] The Design Lab, University of California San Diego, La Jolla, CA, USA.
[3] Department of Computer Science and Engineering, University of California San Diego, La Jolla, CA, USA.
[4] Department of Medicine, Kaiser Permanente, San Diego, CA, USA.
[5] Department of Radiation Oncology, University of California San Francisco, San Francisco, CA, USA
[6] Department of Emergency Medicine, University of California San Francisco, San Francisco, CA, USA
[7] Division of Cardiology, Department of Medicine, University of California San Diego, La Jolla, CA, USA
[8] Department of Orthopaedic Surgery, Korea University Ansan Hospital, Ansan, South Korea
[9] Google Research, Seattle, WA, USA
[10] Paul G. Allen School of Computer Science & Engineering, University of Washington, Seattle, WA, USA.

## Abstract

Online health resources and large language models (LLMs) are increasingly used as a first point of contact for medical decision-making, yet their reliability in healthcare remains limited by low accuracy, lack of transparency, and susceptibility to unverified information. We introduce a proof-of-concept conversational self-triage system that guides LLMs with 100 clinically validated flowcharts from the American Medical Association, providing a structured and auditable framework for patient decision support. The system leverages a multi-agent framework consisting of a retrieval agent, a decision agent, and a chat agent to identify the most relevant flowchart, interpret patient responses, and deliver personalized, patient-friendly recommendations, respectively. Performance was evaluated at scale using synthetic datasets of simulated conversations. The system achieved 95.29% top-3 accuracy in flowchart retrieval (N=2,000) and 99.10% accuracy in flowchart navigation across varied conversational styles and conditions (N=37,200). By combining the flexibility of free-text interaction with the rigor of standardized clinical protocols, this approach demonstrates the feasibility of transparent, accurate, and generalizable AI-assisted self-triage, with potential to support informed patient decision-making while improving healthcare resource utilization.

# Main

Self-triage, the process by which people assess the urgency of their symptoms and decide when and where to seek medical care, is becoming increasingly prevalent with the use of online resources. According to the National Center for Health Statistics, 58.5% of adults searched for health or medical information online during a six-month period from July to December 2022[1]. Popular platforms such as WebMD and Mayo Clinic offer a wide range of health content, including symptom checkers, drug information, and in-depth articles on diseases and conditions. More recently, a growing number of people have begun relying on large-language models (LLMs) for health-related inquiries, complex probabilistic algorithms that generate human-like textual responses, following the introduction of ChatGPT[2] in 2022. A U.S. online survey of 2,002 participants found that 95.6% of the respondents used search engines for health queries, while 32.6% reported using LLMs such as ChatGPT to address medical or health questions in 2024[3]. The trend in self-triage towards LLMs has shifted the initial point of contact in healthcare from clinicians to online platforms, which offer greater accessibility and cost-efficiency, particularly for individuals in low- and middle-income regions with limited healthcare infrastructure.

At the other end of the spectrum, emergency rooms today are frequently overwhelmed with nonurgent visits. Nearly 40% of all emergency department (ED) visits in the U.S. are medically nonurgent[4], which could be more appropriately managed in walk-in clinics or by primary care providers. Such unnecessary ED visits not only result in longer waiting time for patients truly requiring emergency intervention, but also contribute to physician fatigue and burnout. An effective and accessible self-triage tool could enable patients to receive timely triage recommendations and allow them to better manage their conditions and plan for clinic visits. Meanwhile, clinicians can offload the triage burden and enhance the overall efficiency of healthcare resource allocation.

Despite the growing need, developing an informative and reliable self-triage tool remains challenging. While symptom checkers[5–7] can assist with self-triage by providing a list of possible causes or diagnoses, it suffers from low accuracy, requires a certain level of health literacy, and fails to provide definitive actions tailored to specific cases[8,9]. Conversational LLM chatbots such as ChatGPT frequently provide vague or overly general advice and at the same time suffer from the inherent “hallucination problem”, where models fabricate incorrect or misleading information due to their probabilistic prediction nature[10–12]. Moreover, LLMs are considered as a “black box” given their generation process is rarely transparent, making it difficult for users or clinicians to trust or validate the guidance provided[13,14]. Although LLMs can be fine-tuned on domain-specific medical datasets to improve their performance, it is difficult for healthcare professionals to control or audit the output of LLMs before deploying them as medical chatbots for public use. The lack of a shared, structured representation of the medically informed procedures between human experts and LLMs limits their reliability and safety for both patients and providers.

To address these challenges, we propose using clinically validated flowcharts as a shared representation between LLMs and healthcare practitioners for healthcare applications. We demonstrate this concept through a conversational self-triage AI system - TriageMD, explicitly guided by American Medical Association (AMA) flowcharts that are retrieved based on conversation intake. The flowchart format enables the system to guide users through a series of clinically relevant questions, reflecting the established procedural flow clinical nurses use. This approach combines the free-text conversational capabilities of LLMs with the clinical rigor of standardized triage protocols, offering a novel solution for delivering personalized and reliable decision support. We leverage different prompt engineering techniques and a multi-agent framework (Figure 1) to create the system which performs triage tasks through natural conversations. The system is composed of a flowchart database which stores the graph-represented AMA flowcharts (Figure 1b) and three agents powered by LLMs: retrieval agent, decision agent, and chat agent (Figure 1c). These agents collaboratively perform the triage task by retrieving the relevant flowchart based on patient information, interpreting patient responses, and providing patient-friendly guidance, respectively. At the end of the triage process, the system provides final action suggestions (Figure 1a) tailored to each individual case as derived from navigating through the relevant AMA-validated flowchart (Figure 1d).

We evaluated our proof-of-concept system through two tasks: flowchart retrieval, which measures the system's ability to identify the most appropriate flowchart from the database, and flowchart navigation, which assesses its ability to interpret patient responses and provide appropriate step-by-step guidance. Given the intended use in healthcare settings, we aimed to evaluate the system's performance at scale to rigorously test the reliability and faithfulness of its outputs. To this end, we used synthetic datasets generated by different language models for controlled and repeatable evaluation of simulated conversations. The retrieval results show that the system identified the top-3 most relevant flowcharts out of 100 validated AMA flowcharts with an average accuracy of 95.29%. In the navigation task, the system achieved an average accuracy of 99.10% in understanding simulated patient responses with diverse conversational styles (i.e., brief vs descriptive, clear vs vague) and conditions (i.e., yes/no/maybe, off-topic) and guiding conversations towards a final recommendation. These findings support the proof-of-concept system's strong potential for patient-facing decision making with both accuracy and consistency, while maintaining transparency of the LLMs' decision-making process using an explicit, human readable or editable database of medical flowcharts.

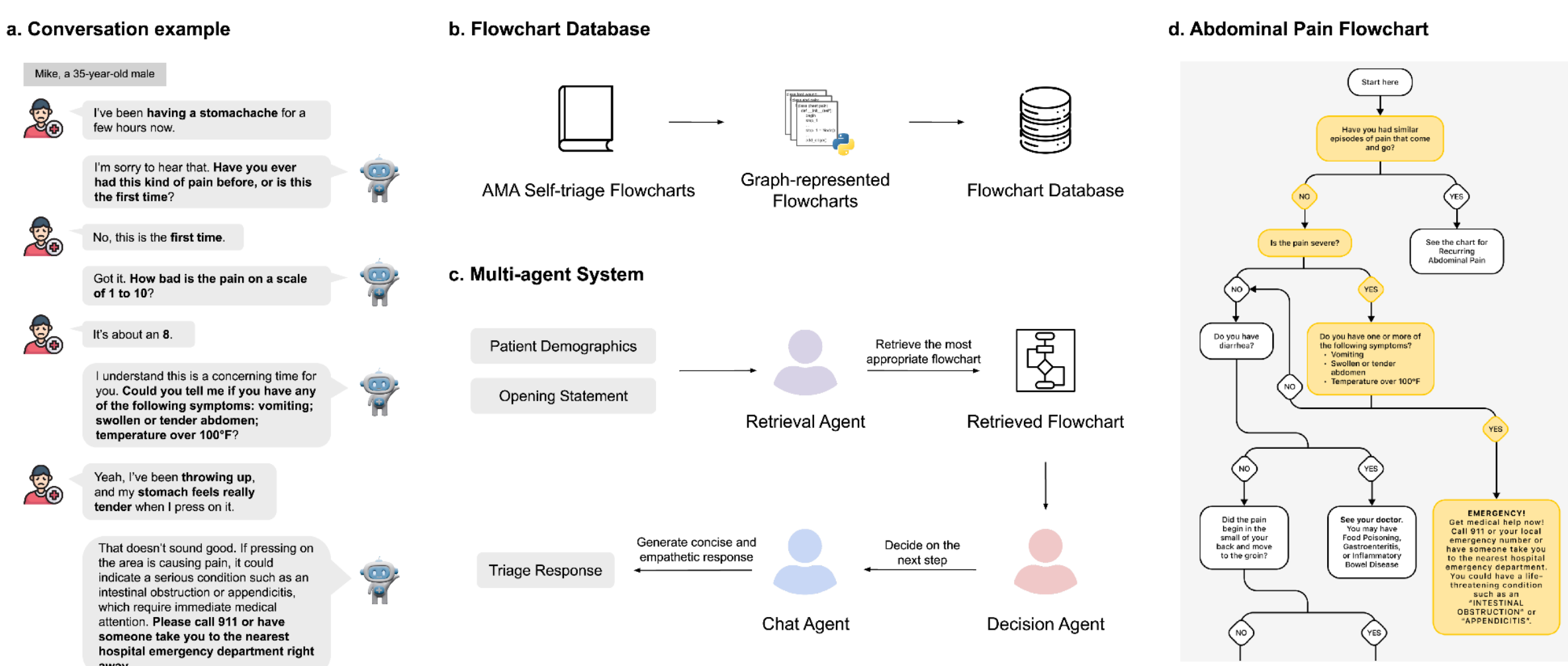

**Fig. 1: Multi-agent self-triage system overview. a.** Example conversation between a 35-year-old male patient and the TriageMD chatbot regarding abdominal pain. **b.** The implementation of flowchart database. We took 100 AMA self-triage flowcharts and converted them to graph representation to create the flowchart database. **c.** Retrieval agent selects the most appropriate flowchart based on patient demographics (35-year-old male) and main concern ("I've been having a stomachache for a few hours now") from the flowchart database. Decision agent interprets patient input and determines the next step based on the flowchart. Chat agent generates concise and empathetic responses to patients, based on both the output of the decision agent. **d.** Example AMA self-triage flowchart for abdominal pain in adults. The AMA flowcharts are structured as a series of polar questions designed to collect necessary clinical information, ultimately guiding the patient towards a recommended triage action, such as self-care, primary care appointment, or emergency evaluation.

# Results

The self-triage system we developed is structured around three key components: a self-triage flowchart database, a retrieval module to identify the most relevant flowchart based on patient information, and a navigation process to interpret and systematically follow the selected flowchart. To assess the system's performance at scale, which is an essential consideration for real-world healthcare assessment, we utilized large synthetic datasets generated by LLMs. In particular, given the back-and-forth nature of conversational patient triaging, it would not be possible to use a pre-recorded or existing dataset of human conversation to test our AI system. We generated synthetic datasets that enable extensive and controlled testing across diverse clinical scenarios, offering both scalability and comprehensiveness which is difficult to achieve through human-subject studies. To enhance generalizability, we utilize multiple LLM models across different companies to create the datasets for system assessment. As an initial proof-of-concept system, our evaluation focuses on the ability of LLMs to accurately retrieve and execute the proper flowcharts. Specifically, we aim to assess two hypotheses:
H1) the retrieval module can accurately identify the most appropriate flowchart from the database;
H2) the navigation module can reliably interpret patient responses and correctly follow the retrieved flowchart.

## Flowchart Database

We first created a flowchart database of self-triage flowcharts derived from the American Medical Association Family Medical Guide, 4th Edition. It covers a wide range of common symptoms and includes flowcharts for particular age or sex groups. We obtained a total of 100 flowcharts after preprocessing (detailed in the Method section) and illustrated the distribution in Figure 2. Notably, the majority of the flowcharts is applicable for all sexes and ages, with subsets targeted toward specific populations, such as female or pediatric. Figure 1b shows the diverse coverage of the flowchart database across medical specialties. 100 flowcharts were categorized into ten specialties, with reproductive health containing the highest number of flowcharts (N=14) and urology being the fewest (N=5).

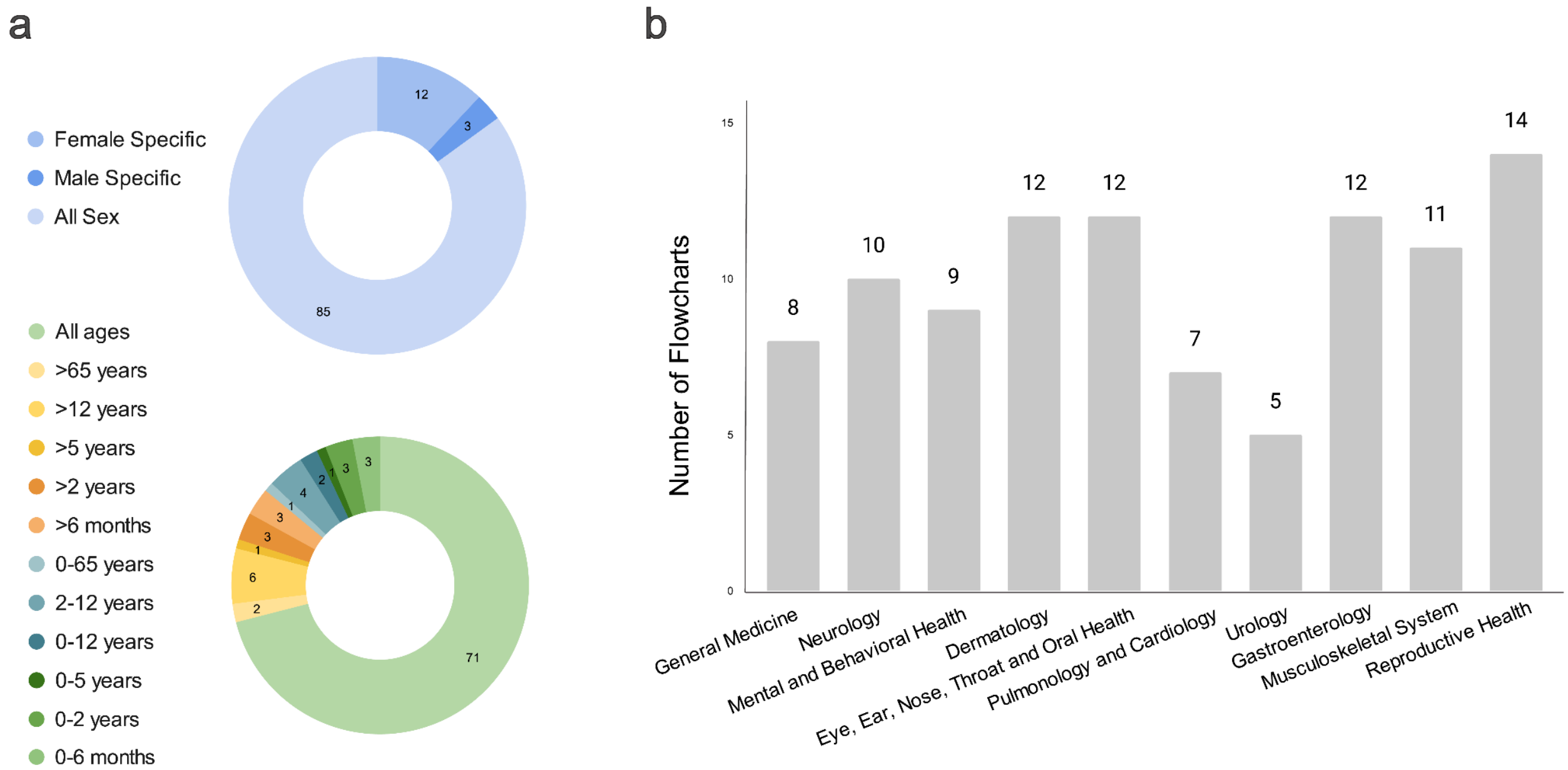

**Fig. 2: Flowchart Database Breakdown. a.** the distribution of flowcharts by target sex and age groups of intended users. **b.** the distribution of flowcharts by medical specialty. A total of 100 flowcharts were categorized into ten specialties, with reproductive health containing the highest number of flowcharts (N=14) and urology the fewest (N=5).

## Retrieving the most appropriate flowchart from the database

We assess the system performance in selecting the most appropriate flowchart from the database based on the patient's demographic information and main concern. This selection is achieved by a retrieval agent implemented using retrieval augmented generation (RAG), as detailed in the Methods section. Synthetic patient demographics and opening statements, defined as the first sentence a patient would use to express their main concern, were generated given the flowchart descriptions using four different models: GPT-4o, Claude 3 Haiku, Gemini 2.0 Flash-Lite, DeepSeek-Chat.

Figure 3 presents the flowchart retrieval result evaluated on the synthetic opening statement dataset. Figure 3a shows the structure of the retrieval agent, which consists of a similarity search module followed by an LLM (we used OpenAI GPT-4o-mini model for cost-efficiency). The patient’s demographic information and opening statement are first processed through a cosine similarity search to retrieve the top-ranked flowcharts along according to their similarity scores. The retrieved candidates are then passed to the LLM, which selects the most appropriate flowchart based on both the flowchart information and the similarity scores. The module components correspond to the retrieval performance in Figure 3b based on N=2000 opening statements comprising 20 per AMA flowchart. When using only the LLM module (Yellow), which relies solely on its semantic comprehension, the average accuracy of correctly identifying the relevant flowchart is 78.40% (± s.d. 6.74%, N=2,000). Similarly, when utilizing only the similarity search module (Green), which mathematically compares distances within vector space, the accuracy of identifying the correct flowchart as the closest is 82.14% (± s.d. 7.10%, N=2,000). When the two modules are combined to form the retrieval agent (Grey), the average accuracy improves to 84.66% (± s.d. 5.59%, N=2,000). This demonstrates that the retrieval agent not only considers the results from the similarity search (with top-n set to 10), but also performs semantic analysis of each retrieved candidate to make a final selection. One common reason for retrieval confusions across all models is that certain symptoms are related and mentioned together in the opening statements (e.g., chest pain and shortness of breath), making it difficult for the retrieval agent to select a single, most appropriate flowchart when multiple candidates exhibit high semantic similarity. We provide a complete failure analysis on the retrieval result in the Supplementary Table 1.

In addition, Figure 3b also presents a comparison of retrieval accuracy across opening statements generated by four different models. In the LLM-only setting, performance on GPT4o and Claude Haiku generated opening statements exhibits similar accuracy: 84.70% and 83.75% respectively, while slightly lower accuracy is observed on Gemini Lite and DeepSeek Chat generated statements at 72.40% and 72.75%. For the similarity search-only setting, accuracy is 86.60% for GPT4o, 89.75% for Claude Haiku, 76.65% for Gemini Lite, and 75.55% for DeepSeek Chat. In the full retrieval agent setting, accuracy reaches 90.60% for GPT4o, 88.25% for Claude Haiku, 79.60% for Gemini Lite, and 80.20% for DeepSeek Chat. A complete result with standard deviation is included in Supplementary Table 5.

Figure 3c specifically evaluates the top results returned from the similarity search component. Here we vary the *n* value to assess the accuracy of whether the correct flowchart appears within the top-*n* results, ranked by similarity score. The top-1 result corresponds to the similarity search performance shown in Figure 3b. When we consider the top-3 retrieved candidates from similarity search, the average accuracy increases to 95.29% (± s.d. 2.88%, N=2,000). Extending this to the top-5 retrieved results further slightly improves the average accuracy to 97.71% (± s.d. 1.52%, N=2,000). These findings suggest that the similarity search module can effectively retrieve and rank the relevant flowcharts. In real-world deployment, we expect an interactive system will present both a “most relevant” flowchart recommendation from the retrieval agent and the top 3 flowchart results from similarity search as “potentially relevant” flowcharts. This retrieval process leverages the AI’s ability to automatically retrieve from a large database while maintaining transparency to allow patients to stay in the loop during the triage process.

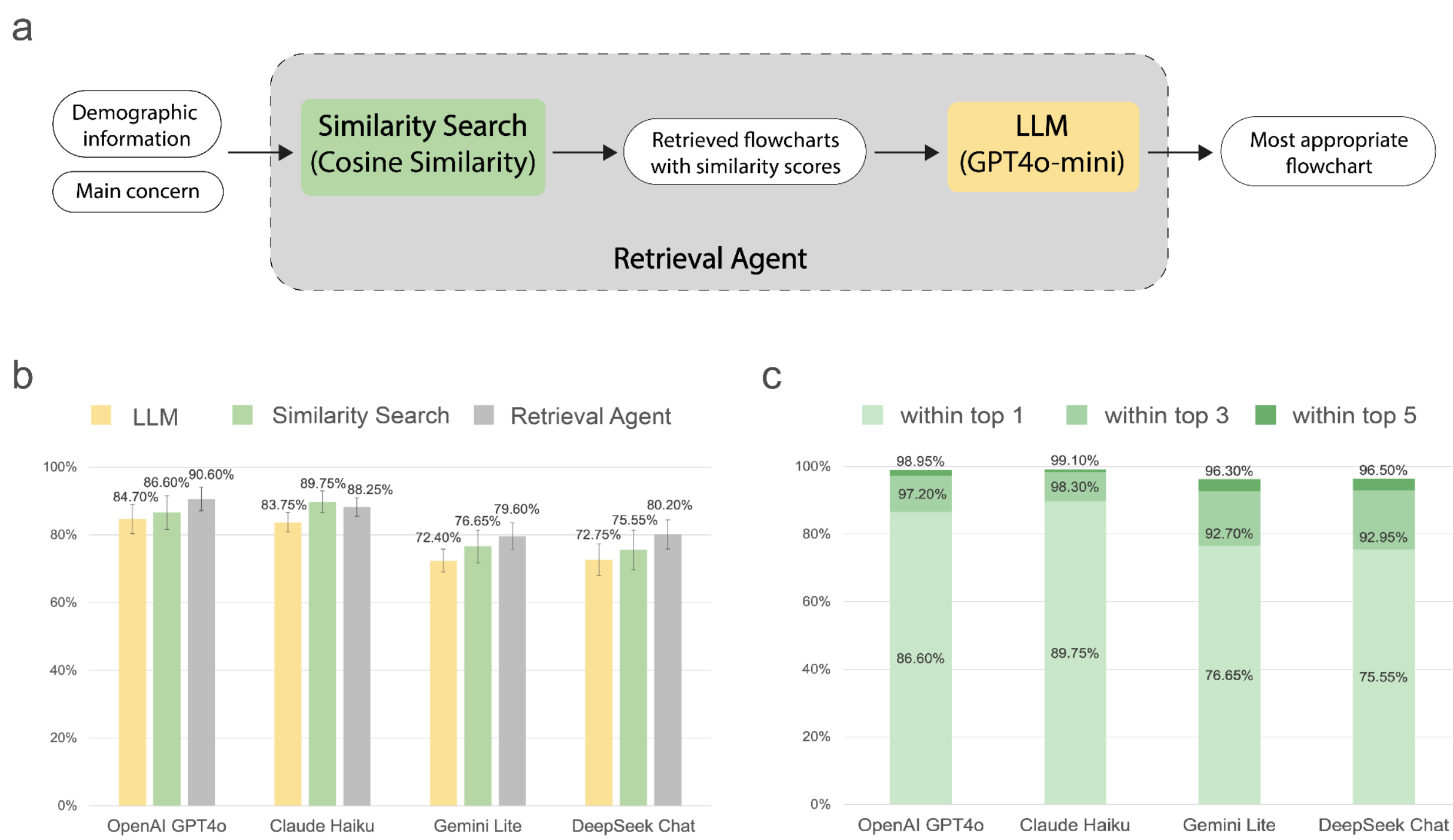

**Fig. 3: Evaluation of flowchart retrieval accuracy on synthetic opening statement data**. **a.** Structure of the retrieval agent, which consists of a similarity search (cosine similarity) module followed by an LLM (GPT-4o-mini). Patient demographic information and opening statements (defined as the first sentence a patient would use to express their main concern) are processed through cosine similarity search to retrieve top-ranked flowcharts with similarity scores, which are then analyzed by the LLM to select the most appropriate flowchart. **b.** Retrieval accuracy across LLM-only, similarity search-only (top-1 result), and combined retrieval agent settings, evaluated on opening statements generated by GPT4o, Claude Haiku, Gemini Lite, and DeepSeek Chat. The retrieval agent achieves the highest average accuracy (84.66% ± s.d. 5.59%, N=2,000) while showing a comparable performance across different LLM-generated statements. Note in retrieval agent setting, similarity search module returns top-10 flowchart to be considered by the LLM module. **c.** Retrieval accuracy from top 1, top 3, and top 5 retrieved results. We further extended the similarity search results to the top 3 and 5. The average accuracies for retrieving the correct flowchart within the top 3 and 5 retrieved results are 95.29% ± s.d. 2.88% and 97.71% ± s.d. 1.52%, respectively (N=2,000). This suggests that the similarity search module can effectively retrieve and rank the relevant flowcharts, which supports a more inclusive decision-making process.

## Navigating through the retrieved flowchart

To provide proper guidance to patients, it also requires the system to correctly navigate through the retrieved flowchart in a step-by-step conversational manner. We further evaluated the system's navigation performance using synthetic patient responses generated by the same four language models as aforementioned. Drawing from real-world patient-doctor interaction data[15–17], we categorized patient responses to doctors' queries into five conversational patterns (See Methods section for detailed description and response examples):

1) brief: responses that are short and conclusive, suggesting a clear *yes* or *no*;
2) descriptive: responses that are descriptive and conclusive, providing additional details;
3) weak: responses that are vague or ambiguous, but slightly leaning towards *yes* or *no*;
4) uncertain: responses that are too inconclusive to be interpreted as a *yes* or *no*;
5) off-topic: responses that are completely irrelevant to the question.

These patterns represent diverse patient response styles, while preserving conversational realism. We created a large-scale response dataset that includes all five patterns, resulting in a total of 37,200 synthetic patient responses across 100 flowcharts (see Methods section for generation details). To assess the navigation performance, the patient response and flowchart node pairs are fed into the decision agent. The decision agent is designed to classify patient responses along four axes: On-topic/Off-topic, Answered/Unanswered, Yes/No, Certain/Uncertain. These axes collectively determine whether the system can confidently advance to the next node or needs to seek clarification, thus enabling a transparent and informative navigation process.

Figure 4 presents the navigation result based on the decision agent's output (the decision agent was implemented with OpenAI GPT-4o-mini model as the backbone). Overall, the system demonstrates high navigation accuracy across patient responses generated by all four models, achieving an average of 99.10% (± s.d. 0.54%, N=37,200), as shown in Figure 4a. Figure 4b-4f provide a detailed breakdown of performance across five conversational patterns. For patterns 1-3, we further break down the result into four categories: a: Certain and Correct (light green); b: Uncertain and Correct (green); c: Uncertain and Incorrect (blue-green); and d: Certain and Incorrect (red). The system only proceeds when it determines the response is certain and holds at the current position if uncertainty is detected. As such, the Certain and Incorrect category is the only unacceptable case, as the system misinterprets the patient response and proceeds with a potential of providing false guidance to patients, whereas category a, b, and c are all considered acceptable.

For pattern 1 (brief responses), the system achieved an average accuracy of 98.11% ± s.d. 2.13% (N=9,300) across all LLM-generated responses for acceptable decisions (a: 96.88% ± s.d. 2.04%; b: 0.86% ± s.d. 0.50%; c: 0.37% ± s.d. 0.26%), with 1.89% ± s.d. 2.13% falling into the unacceptable category (d), as shown in Figure 4b. For pattern 2 (descriptive responses), the average accuracy for acceptable decisions is 99.65% ± s.d. 0.14% (a: 96.85% ± s.d. 1.06%; b: 2.58% ± s.d. 1.22%; c: 0.21% ± s.d. 0.27%, N=9,300), with only 0.35% ± s.d. 0.14% classified as Certain and Incorrect (d), as illustrated in Figure 4c. For pattern 3 (weak responses), the system maintained high performance with 99.69% ± s.d. 0.22% (N=9,300) acceptable decisions (a: 23.49% ± s.d. 11.85%; b: 55.38% ± s.d. 10.82%; c: 20.81% ± s.d. 1.89%), and 0.31% ± s.d. 0.22% unacceptable (d) in Figure 4d. This result suggests that even when patient responses were intentionally ambiguous, the TriageMD system was able to detect the underlying uncertainty, as a large portion of the responses is classified as uncertain (category b and c).

Since both pattern 4 (uncertain responses) and pattern 5 (off-topic responses) are designed to be inconclusive, the correct system behavior is to refrain from proceeding to the next node until a certain, on-topic response is provided. For pattern 4, the navigation process is considered correct if it remains at the current step and seeks confirmation from the patient. The performance for pattern 4 across responses generated by four language models is presented in Figure 4e. Similarly, the acceptable cases are: a: Certain and Unanswered; b: Uncertain and Unanswered; c: Uncertain and Answered, and the only unacceptable case is d: Certain and Answered. The average accuracy for acceptable decisions is 99.53% ± s.d. 0.49% (a: 0.45% ± s.d. 0.15%; b: 85.98% ± s.d. 11.56%; c: 13.10% ± s.d. 10.96%, N=4,650), with only 0.47% ± s.d. 0.49% classified as unacceptable (d).

For pattern 5, the navigation process is considered correct if the system classifies the response as off-topic and restates the question. As shown in Figure 4f, the average accuracy across all LLM-generated patient responses is 98.26% ± s.d. 3.39% (N=4,650), with the highest error rate being 6.82% with Claude Haiku. Overall, the system exhibits high accuracy on flowchart navigation across patient responses generated by different models. It effectively handles varying degrees of descriptiveness, ambiguity, and irrelevance, consistently producing appropriate guidance and maintaining the integrity of the triage protocol. We provide a complete failure analysis for flowchart navigation results in Supplementary Table 3 and a complete evaluation results with standard deviation in Supplementary Table 6.

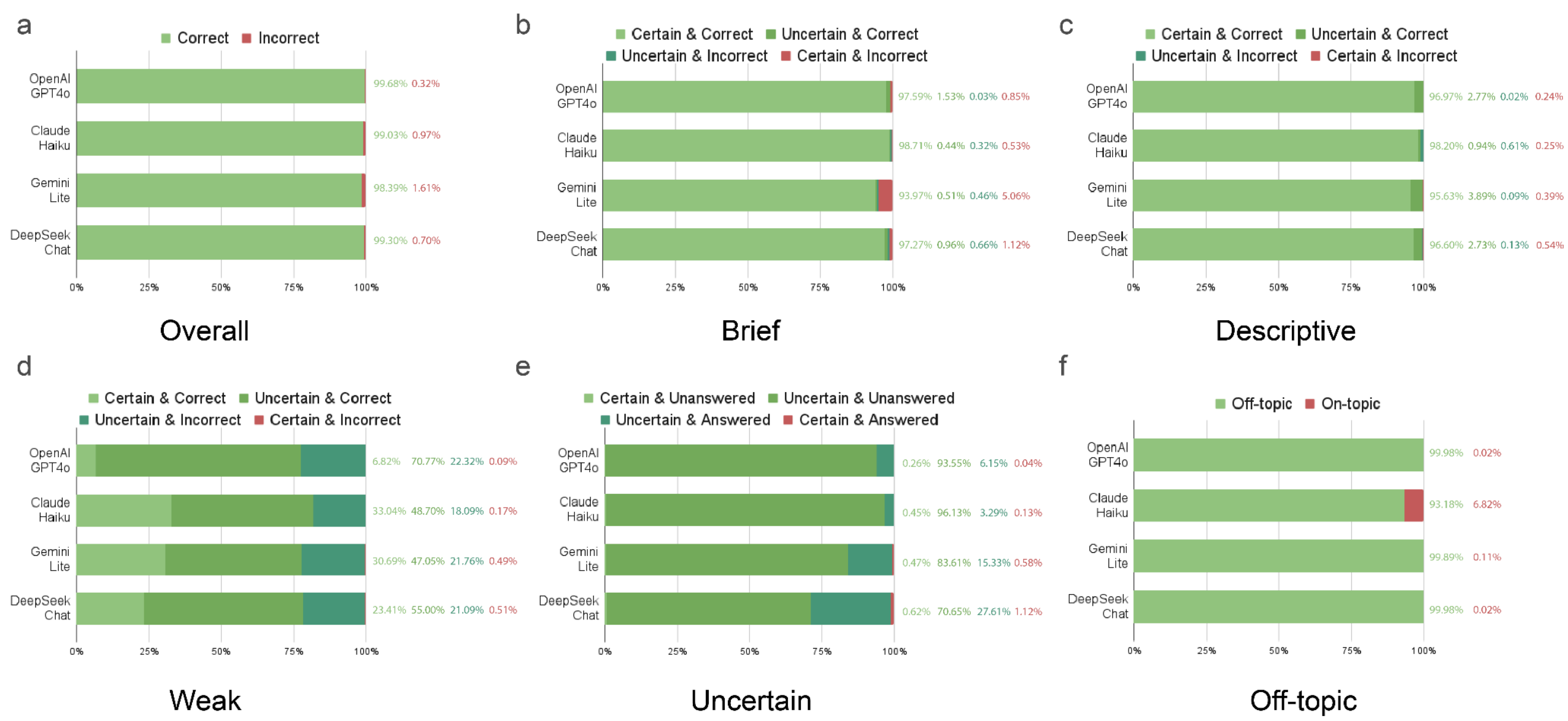

**Fig. 4: Evaluation of flowchart navigation accuracy on synthetic patient response data. a.** Overall accuracy across patient responses generated by four models. The system achieves 99.10% on average (± s.d. 0.54%, N=37,200). **b-f.** Navigation performance breakdown across five conversational patterns: brief, descriptive, weak, uncertain, and off-topic (see Methods for detailed pattern explanations and examples). For the first three patterns (**b-d**), we present the results in four categories: Certain and Correct (light green), Uncertain and Correct (green), Uncertain and Incorrect (blue-green), and Certain and Incorrect (red). Only the last category is unacceptable since the system would not proceed if uncertainty is detected. Acceptable accuracy reached 98.11% ± 2.13% (n = 9,300) for brief responses and 99.65% ± 0.14% (n = 9,300) for descriptive responses. For weak responses that are designed to be ambiguous, the acceptable accuracy was 99.69% ± s.d. 0.22% (N=9,300) on average, with the majority classified as uncertain. For uncertain responses (**e**), we slightly modified the category name to reflect their non-answer nature: Certain and Unanswered (light green), Uncertain and Unanswered (green), Uncertain and Answered (blue-green), and Certain and Answered (red). The system achieved 99.53% ± s.d. 0.49% (N=4,650) accuracy for acceptable categories. Off-topic responses (**f**) were correctly identified at 98.26% ± 3.39% (n = 4,650). The system demonstrates robust navigation accuracy across diverse conversational patterns.

# Discussion

We introduce a multi-agent self-triage system to demonstrate how clinically validated flowcharts can effectively guide LLM outputs in healthcare applications. Self-triage is a complex procedure that involves information gathering, decision making, and conversation comprehension, while adhering to the high fidelity required in healthcare settings. Leveraging self-triage flowcharts developed by AMA, we designed three LLM-powered agents to procedurally perform the triage task through natural conversations: retrieval agent, decision agent, and chat agent. To evaluate system performance, we conducted two key assessments: flowchart retrieval and flowchart navigation. These evaluations measure the system's ability to accurately select the appropriate flowchart based on patient input and correctly navigate through the flowchart to provide guidance. The results show that the system successfully performs the end-to-end triage workflow with both high accuracy and clinical alignment.

Our results have multiple implications to real-world deployment. First, one of the key contributions of our system is its ability to perform triage strictly according to clearly defined, auditable medical protocols. By leveraging flowcharts as the shared representation, clinicians can review the exact decision path taken during each conversation session, providing a layer of transparency and accountability that is often missing in black-box AI systems[18–20]. Moreover, this design can empower clinical systems/teams to customize the triage workflow to fit their practice by editing or modifying the flowcharts. In the future, human-AI collaboration could be employed to expand the flowchart database to cover a wide range of clinical tasks: LLMs can propose new flowcharts based on real-world data, while humans serve as expert reviewers to validate and refine them prior to deployment.

Second, our flowchart retrieval results suggest that in real-world deployment, both the top 3 flowchart results and the final flowchart recommendation from the retrieval agent should be presented to patients (Figure 3). While the retrieval agent achieved the highest accuracy (84.66%) in Figure 3b, it could be complemented by the top 3 retrieved results from similarity search (95.29%) as shown in Figure 3c. Presenting both options allows flexibility in patient interaction: the system can offer a clear recommendation for those seeking guidance, while also enabling patients who prefer to be more actively involved to compare alternatives. This approach supports both system accuracy and greater transparency in the triage process.

Third, our flowchart navigation results highlight that a key feature for patient-facing conversational systems is the ability to handle a wide range of patient response styles, particularly those involving uncertainty. In clinical practice, communication skills are a fundamental part of clinician training[21,22]. During patient interactions, clinicians need to extract essential information from varied patient responses, make diagnostic and treatment decisions, and communicate those decisions clearly. Among the five patient response patterns we defined, brief responses appear more frequently when physicians pose yes-no questions, while weak or uncertain responses are also common in patient-doctor interactions and present greater challenges for interpretation[23,24]. This is reflected in Figure 4d-4e, where a large proportion of responses fall into the uncertain categories. For real-world deployment, the system should be designed to proceed only when a confirmed answer is received, thereby reducing the risk of providing inappropriate action suggestions. Future improvements can also be made on flowchart design. For example, assigning weights to each question in the flowcharts could enable the system to prioritize high-value nodes and adjust its navigation dynamically. Additionally, conditional paths for uncertain responses could be added to flowcharts to address uncertainty more explicitly. Each yes-no question node could have an additional path to specify the desired action when the response is uncertain, such as merging with one of the existing paths, triggering an alternative question, or referring the patient to a clinician.

Fourth, our results also show minor variation in performance across synthetic data generated by different language models. In Figure 3b, we observed that compared to OpenAI GPT4o and Claude Haiku, the retrieval accuracy on opening statements generated by Gemini Lite and DeepSeek Chat was lower across all settings (LLM-only, Similarity Search-only, and the full retrieval agent) in Figure 3b. One possible explanation is that given the same prompt for generation, the synthetic opening statements from Gemini Lite and Deepseek Chat tend to be more generic, leading to a broader and less targeted list of candidate flowcharts during the similarity search stage. For navigation results, Gemini Lite generated responses exhibited slightly higher error rate on pattern 1 (Figure 4b), while Claude Haiku generated responses presented a relatively higher error rate on pattern 5 (Figure 4f). Manual inspection revealed that Gemini Lite often produces less direct affirmative responses when prompted for brief and affirmative responses. When instructed to generate off-topic responses, Claude Haiku tends to first provide an on-topic answer and then introduce unrelated content, a characteristic linked to its design as a helpful AI assistant.

Despite promising initial results, our system has certain limitations. First, the current set of flowcharts is relatively generic and restricted to polar (yes/no) questions, which limits the system's expressiveness and clinical breadth. In the future, more customized and detailed flowcharts could be developed for specific specialities, institutions, or patient populations. Second, our evaluation was conducted using synthetic datasets for extensive assessment. Real-world evaluation with patients is necessary to validate effectiveness and safety of the system. Third, the current system does not yet include mechanisms to detect or recover from errors in retrieval or navigation. Future work could focus on developing error-handling strategies, such as confidence estimation or clinician-in-the-loop interventions. Fourth, privacy also remains important, as users may share sensitive demographic information and medical history during conversations. Addressing these risks through secure data management and transparent consent mechanisms would be critical. Finally, we only developed and tested the system in English with text input; further improvement could expand to multilingual or even multimodal input. For example, patients can upload images to help describe their symptoms and a visual-language model could be developed to perform the triage task.

In conclusion, we propose medically validated flowcharts as a shared presentation between healthcare practitioners and LLMs, and present a conversational self-triage system with AMA flowcharts. Our findings suggest that LLMs have great potential to serve as the front-line triage tools which would offer great benefits to both patients and healthcare providers. While further validation on real-world data is required, the framework lays the groundwork for safe, auditable, and accessible decision support tools that are applicable for a wide range of clinical workflows. By prioritizing transparency, adaptability and efficacy, this work shines a light on a new direction for leveraging structured clinical knowledge to implement explainable and trustworthy AI-driven health tools.

# Methods

## Self-triage Flowcharts and Preprocess

We analyzed and preprocessed all flowcharts from the American Medical Association Family Medical Guide, 4th Edition, published in 2011 by the American Medical Association (AMA). This guide offers general information on common medical conditions, self-diagnosis flowcharts, and actionable health recommendations, aiming to equip individuals and families with essential medical knowledge about their well-being. The self-diagnosis flowcharts follow a question-answering format, focusing on providing triage actions while suggesting possible diagnoses.

To construct our flowchart database, we implemented a series of preprocessing steps. First, we divided the original Infertility flowchart into two separate flowcharts: "Infertility in Men" and "Infertility in Women", since the original flowchart contains two entry points for these two population groups. Second, we modified the flowchart content to fit the goal of self-triage. While self-diagnosis attempts to identify specific medical conditions, self-triage emphasizes providing appropriate actions based on the urgency of symptoms. Our modifications focus on ensuring that each flowchart endpoint delivers a clear, actionable recommendation. This included removing the page references that redirect users to other sections of the book after action items and replacing the term "self-diagnosis" with "self-triage" to better reflect the role and goal of the system. An example of flowchart modification is included in the Supplementary Figure 1. Third, there are three flowcharts that could not be retrieved directly, as their descriptions redirect users to another chart. We mapped them to their respective parent flowcharts to preserve the logical structure and enable future retrieval.

## Flowchart Representation

Given the decision-tree structure of the AMA flowcharts and prior research indicating that LLMs exhibit better comprehension of code-based formats[25–27], we converted the flowcharts into directional graphs using Python. This graph-based representation allows LLMs to more effectively parse the flow of decision-making, enabling accurate step-by-step reasoning and execution. In our graph representation, each step in the flowchart is modeled as a node, and transitions between steps are directed edges. We defined four node types to capture the functional semantics of the flowcharts: 1) N node, which poses yes-or-no questions to the user and is the most common node type; 2) F node, which denotes nodes that redirect users to a linked flowchart; 3) A node, which represents the endpoints of the flowchart, typically an action recommendation (e.g., "See your doctor") or an educational message (e.g., "Headaches often occur with fevers."); and 4) I node, which provides brief explanations or reassurance preceding an F node. Notably, nodes are connected by directional and conditional edges, where the direction points to the next step, and the condition is either Yes or No as the answer to the N node. An example of the graph representation is provided in the Supplementary Figure 1.

## Multi-agent Self-triage System

Here we introduce the multi-agent system, which consists of three LLM-based agents working sequentially to perform self-triage: the retrieval agent, the decision agent, and the chat agent. The system first collects patient information, including sex, age, and main concern. This information is utilized to identify the most relevant flowchart from the database by the retrieval agent. After finalizing on a flowchart, the decision agent and the chat agent collaborate to execute the triage protocol. Specifically, the decision agent interprets and categorizes the patient's answer to the flowchart question, while the chat agent generates empathetic and concise responses based on the flowchart content. This modular architecture is designed to reflect clinical decision-making workflows, with each agent playing a specialized role that mirrors how healthcare professionals gather information, interpret responses, and communicate care recommendations.

### Patient Information Gathering

The triage process begins by collecting the basic demographic information of patients, including sex (female/male) and age, with age units specified in months or years, given the database contains pediatric flowcharts for infants under one year old. Following this, the conversation is initiated by TriageMD prompting the patient to state their main concern (also referred to as chief complaint in clinical settings). For example, a patient might begin with "I've been having a stomachache for a few hours now", as shown in Figure 1. The demographic details and the patient's opening statement are then combined and provided to the retrieval agent, which requires this input to identify the most appropriate flowchart from the database.

### Retrieval agent - find the most appropriate flowchart via similarity search

We leverage the retrieval-augmented generation (RAG) framework to effectively match patient concerns with the most appropriate self-triage flowcharts. To build the retrieval component, we first curated the flowchart descriptions from the source book and converted them into embeddings using Facebook AI Similarity Search (FAISS). Upon receiving a patient's demographic details and main concern, this information is also vectorized into an embedded query. We then compare it against the embedded flowchart descriptions to identify the top-n most relevant candidates using cosine similarity. These candidate flowcharts, along with the patient's query, are then provided to a LLM, which selects the final, most appropriate flowchart. If no suitable flowchart matches the patient's case, “no flowchart available” would be returned, and the system would prompt the patient to seek medical attention directly as it is not authorized to provide guidance for conditions not covered by the flowchart database. The detailed prompt for creating the retrieval agent is included in the Supplementary Table 7.

**Decision agent - interpret patient's responses**

After retrieving the appropriate flowchart, the system begins following it. This involves interpreting patient responses, determining whether patients have answered the flowchart question, and navigating through the flowchart to identify the next step. Specifically, at each question node, the decision agent takes in both the node content and the patient's responses as input and outputs answers to four different axes detailed in Table 1. These axes and their descriptions are included in the agent prompt using a structured output parser, which ensures that the output of the decision agent follows a well-defined, structured format (JSON). For example, in the abdominal pain flowchart, the first node asks “Have you had similar episodes of pain that come and go?” If the patient responds, “No, this is the first time”, the decision agent generates structured output as follows: isOnTopic=Yes, isAnswered=Yes, actualAnswer=No, and isUncertaiN=No. The detailed prompt is included in the Supplementary Table 7. Each axis is essential for determining the next step: the system proceeds to the subsequent node only when a certain (isUncertaiN=No) yes or no answer (actualAnswer=Yes or actualAnswer=No) is obtained. Otherwise, it remains at the current node and attempts to clarify (isOnTopic=No) or confirm (isUncertaiN=Yes) the patient's response. Decision agent plays a critical role in the system for navigating through flowchart nodes and sending the correct content to the next agent - chat agent.

| Axis Name | Description |
|---|---|
| isOnTopic | Return “Yes” if the response is relevant to the question, otherwise return “No”. |
| isAnswered | Return “Yes” if the response provides a yes or no answer to the question, otherwise return “No”. |
| actualAnswer | Return “Yes” if the patient answers the question affirmatively, “No” if the patient answers negatively. |
| isUncertain | Return “Yes” if the response expresses uncertainty (e.g., 'maybe', 'not sure', 'probably'), otherwise return “No”. |

**Table 1** Four axes defined for the decision agent's structured output

**Chat Agent - provide concise and empathetic responses to patients**

The chat agent serves as the final module in the system, responsible for delivering concise and empathetic responses to the patients. It receives the proper node content from the decision agent and generates a reply tailored to the context of the conversation. The chat agent's prompt is dynamically constructed based on the output of the decision agent, reflecting the patient's previous response. When the patient clearly answers the question, the chat agent simply conveys the flowchart node content in a patient-friendly manner. If the response is off-topic, the chat agent gently acknowledges the patient's input and redirects the conversation by restating the original question. In cases where the response indicates uncertainty, the chat agent attempts to confirm the answer with the patient. This design allows the agent to maintain coherence over multiple dialogue turns while adapting its tone and messaging to the patient's needs.

## Evaluation Methods

**Flowchart Retrieval**

In order to evaluate the retrieval accuracy of the flowchart retrieval module, we first created a synthetic patient opening statement dataset using multiple language models. An opening statement is defined as the first sentence a patient would use to express their main concern to a doctor. This statement, along with demographic information, serves as the input for identifying the most appropriate flowchart. For each flowchart, we generated opening statements by providing its flowchart name, description, applicable age range, and targeted sex group to the models. The flowchart name was later used as the ground truth label for assessing retrieval accuracy.

We selected four language models from different providers for a comprehensive analysis: GPT-4o from OpenAI, Claude 3 Haiku from Anthropic, Gemini 2.0 Flash-Lite from Google, and DeepSeek-Chat from DeepSeek, mainly considering the cost efficiency. Each model generated 20 opening statements per flowchart: 10 brief statements (≤25 words), 10 descriptive statements (≥50 words) to reflect a wide range of real-world patient communication styles. In total, 8,000 opening statements were generated for use in the evaluation. We include the prompt for generating patient demographics and opening statements in Supplementary Table 7, along with generated samples from each model in the Supplementary Table 2.

Post-generation processing was applied to standardize the demographic information and opening statements, ensuring they were formatted correctly and free of extraneous content. Notably, for flowcharts targeting pediatric populations (under 12 years old), the statements were phrased from a caregiver's perspective (e.g., "My daughter keeps coughing at night. What should I do?"). Each opening statement, along with its associated demographic data, was input into the retrieval agent, which returned the most appropriate flowchart. The retrieved result was then compared to the ground truth to compute retrieval accuracy. We evaluated the sub-component of the retrieval agent both individually and combined, which demonstrates that the complete RAG framework improves flowchart retrieval accuracy, offering greater control and transparency in the triage process.

### Flowchart Navigation

Navigating through the flowcharts according to the patient's responses is another key procedure in our triage system. Similarly, we first created a patient response dataset using the same models as stated in the above section. In order to better simulate the real-world conversational patterns from patient-doctor interactions, we proposed five patterns: 1) brief; 2) descriptive; 3) weak; 4) uncertain; 5) off-topic. Detailed explanation along with response examples for each pattern is shown in Table 2.

| Pattern | Definition | Characteristics | Example |
|---|---|---|---|
| **Brief** | **Conclusive and minimalistic:** Responses that clearly answer the question without additional reasoning, details, or repetition of the question. | 1. Brief and direct<br>2. No elaboration or rephrasing the question.<br>3. Focus on “Yes” or “No” | Q: “Are you more than three months pregnant?”<br>A: “Nope.” |
| **Descriptive** | **Conclusive and descriptive:** Responses that clearly answer the question and provide additional details, context, or elaboration to support the answer. | 1. Clear and definitive<br>2. Expand beyond simply answering the question.<br>3. Always include details, context, or a personal anecdote. | Q: “Are you more than three months pregnant?”<br>A: “No, I just had a test recently, and it’s negative.” |
| **Weak** | **Vague or partially conclusive:** Responses that lean towards an answer but include uncertainty or hedge the statement with ambiguous language. | 1. Show doubt or ambiguity.<br>2. Provide a partial inclination toward “Yes” or “No” but doesn’t fully commit.<br>3. Indicators like “I guess”, “Maybe”, “Possibly” | Q: “Are you more than three months pregnant?”<br>A: “I doubt it, but I guess it’s possible” |
| **Uncertain** | **Inconclusive:** Responses that remain uncertain due to a lack of sufficient information, neither confirming nor denying the question. | 1. Uncertain.<br>2. Indicators like “I don’t know” or “I’m not sure”. | Q: “Are you more than three months pregnant?”<br>A: “I’m not sure. I haven’t checked yet.” |
| **Off-topic** | **Irrelevant:** Responses that are completely unrelated to the question but still make basic conversational sense. | 1. Off-topic but plausible.<br>2. Introduces unrelated information that does not pertain to the question. | Q: “Are you more than three months pregnant?”<br>A: “Oh, I’ve been organizing my closet lately. It’s such a mess!” |

**Table 2** Five different conversational patterns defined for synthetic patient response generation.

For each question node in the flowcharts, we generated 5 patient responses per conversational pattern. Since the conclusive and partially conclusive patterns require a Yes or No answer, five Yes and five No responses were generated for each of the first three patterns listed in Table 2. This results in 40 patient responses per question node. Given that 930 question nodes exist across 100 flowcharts, this resulted in 37,200 patient responses per model and an aggregate of 148,800 responses across the four models included in the dataset. During response generation, the prompt included the flowchart question, specified conversational pattern, its definition, and a predefined answer if applicable. A complete version of the prompt is available in the Supplementary Table 7 along with the generated samples in Supplementary Table 4. These generated responses were then input into the decision agent to evaluate flowchart navigation performance. From the four-axis output provided by the decision agent, we further computed the accuracy by evaluating whether the suggested action from the decision agent is correct or not. An incorrect case was defined as one in which the decision agent misinterpreted patient responses and led to inappropriate guidance. This evaluation task measures the robustness of the flowchart navigation process in a controlled and large-scale setting, offering valuable insights into how patient-chatbot interactions can be effectively guided by structured clinical protocol.

## Data availability

The American Medical Association self-triage flowcharts used in this work are published and can be accessed by obtaining the original volume. Owing to copyright restrictions, the complete set of flowcharts cannot be redistributed by the authors. We provide a representative example in the supplementary materials to illustrate the general structure and format of the flowcharts. The full synthetic datasets generated for evaluation are available from the corresponding author upon reasonable request. We provide the prompts for reproducing the datasets and samples from the datasets in supplementary as well.

## Code availability

The code for system implementation and evaluation will be made available on Github upon acceptance of the paper.

## Acknowledgements

The authors thank Nadir Weibel and Matin Yarmand for their valuable feedback during the scoping and implementation stages of this work. We would like to acknowledge Mastafa Springston, Zi Lin, and Robert MacAulay for their expert consultation. We also thank Fuling Sun, Zhihan Zhang, and Varun K. Viswanath for their constructive feedback on the evaluation tasks, figures, and manuscript.

## Author contributions

Y.L., G.C., L.Q, X.L, E.J.W conceived the work. Y.L., S.Y., H.J contributed to the technical implementation. J.W., A.Q., T.L., M.R contributed to the clinical insights of this work. Y.L. and S.Y. drafted the original manuscript. E.J.W supervised the research. All authors reviewed and approved the manuscript.

## Ethics declarations

**Competing interests** The authors declare no competing interests.

# Supplementary Materials

**Table 1: Failure analysis for flowchart retrieval evaluation task**

| Model | Failure Type | Failure Definition | Occurrence Ranking | Failure Example | | | | | | |
|---|---|---|---|---|---|---|---|---|---|---|
| | | | | Pattern | Demographics | Opening Statement | Correct Flowchart | Retrieval Output | Explanation of Failure | Correct flowchart listed in Top5 retrieved results? |
| OpenAI GPT4o | Symptom Overlap | The opening statement generated for one flowchart contains similar keywords or symptoms that correlate with another flowchart. | 1 (50.0%) | Brief | Male, 50 years | There are itchy, discolored patches on my abdomen that have been there for a while. | Itchy Spots And Rashes Flowchart | General Skin Problems Flowchart | The user mentions itchy, discolored patches, but these unusual patches can also be considered symptoms for a general skin problem, which is why the agent chooses the wrong flowchart. | ☑ |
| | | | | Detailed | Female, 47 years | Recently, I've had this persistent ache on the left side of my forehead that just won't go away. It feels like a dull pressure, especially when I lean forward or bend down. I've tried over-the-counter painkillers, but they don't seem to help much. I'm beginning to worry it might be something serious. | Pain In The Face Flowchart | Headache Flowchart | The user mentions aching on the forehead. Since it lies in the head region as well, the agent identifies the condition as a headache. | ☑ |
| | Profile-Influenced | Because of the patient's profile, the system picked a more age-specific or gender-specific flowchart. | 2 (31.4%) | Brief | Female, 3 months | My baby seems fussier than usual and just not herself. | Feeling Generally Ill Flowchart | Crying In Infants Flowchart | The user mentions that their baby is behaving abnormally. This could be classified as feeling generally ill, but because the patient is an infant, the agent overemphasizes the baby's fusiness. | ☑ |
| | | | | Detailed | Male, 75 years | Hello, I've noticed some changes in my behavior that are concerning. I've been more forgetful and confused lately, even with simple tasks. This is not typical for me, and I'm worried it might be a sign of something serious. What should I do next? | Unusual Behavior Flowchart | Confusion In Older People Flowchart | The user mentions that they've noticed some behavior that is not typica. However, because this patient is older, the agent identified this unusual behavior to be caused by confusion. | ☐ |
| | Confusing Opening | The opening includes details that distract from the main symptom or condition. As a result, the model may misidentify the user's intent, linking the case to another flowchart or returning "no flowchart available." | 3 (18.6%) | Brief | Female, 29 years | I'm experiencing strange sensations, like bugs crawling on my skin. | Hallucinations Flowchart | Numbness Or Tingling Flowchart | The user mentions a feeling of bug crawling on their skin, which suggests kind of hallucination. However, the physical framing of the symptoms causes the agent to misidentify as numbing or tingling. | ☐ |
| | | | | Detailed | Male, 28 years | Hi, I'm a 28-year-old male and have been experiencing redness and swelling around my eyes and cheeks for the past few days. It's quite uncomfortable and seems to get worse throughout the day. I haven't changed any of my skincare routines. Could this be an allergic reaction? | Facial Skin Problems Flowchart | no flowchart available | The user mentions facial skin problems, such as redness and swelling. However, the user also addresses their skincare routine and the possibility of an allergic reaction, which shift the focus and lead to misidentification. | ☑ |
| Claude Haiku | Symptom Overlap | The opening statement generated for one flowchart contains similar keywords or symptoms that correlate with another flowchart. | 1 (42.7%) | Brief | Female, 17 years | I've been experiencing really painful and heavy periods lately. Is this normal, or should I see a doctor? | Painful Periods Flowchart | Heavy Periods Flowchart | The user mentions having painful and heavy periods. Both of these adjectives are relevant to "Painful Periods Flowchart" and "Heavy Periods Flowchart," so the agent misidentifies. | ☑ |
| | | | | Detailed | Male, 54 years | I've been dealing with these strange, hard lumps that have been popping up all over my body. They're not painful, but they're definitely noticeable and a bit concerning. I'm not sure if they're related to any underlying health issues, and I'm wondering if I should come in to have them checked out. | Raised Spots And Lumps Flowchart | Swellings Under The Skin Flowchart | The user mentions lumps forming on their body. Lumps on the body can be related to swellings under the skin as well, so the agent misidentifies. | ☑ |
| | Profile-Influenced | Because of the patient's profile, the system picked a more age-specific or gender-specific flowchart. | 2 (40.2%) | Brief | Male, 75 years | I've been experiencing a general sense of not feeling quite right, and I'm concerned about my health. Can you provide some guidance on what I should do? | Feeling Generally Ill Flowchart | Confusion In Older People Flowchart | The user mentions feeling generally unusal. However, because of the user's age, the agent identifies this as confusion in an older person. | ☑ |
| | | | | Detailed | Male, 5 years | My son has been complaining of pain in his right shoulder for the past few days. He says it hurts when he tries to lift his arm or move it around. He's also been having trouble sleeping at night because the pain seems to be worse then. I'm worried that he may have injured himself playing at the park or during his soccer practice. He's usually very active, but this is really slowing him down. Can you give me some advice on what I should do to help him feel better? | Painful Shoulder Flowchart | Limping In Children Flowchart | The user mentions their son having pain in his shoulder, which is slowing him down. However, because of the age provided, the agent chose "Limping in Children Flowchart" as the final flowchart. | ☑ |
| | Confusing Opening | The opening includes details that distract from the main symptom or condition. As a result, the model may misidentify the user's intent, linking the case to another flowchart or returning "no flowchart available." | 3 (17.1%) | Brief | Male, 68 years | I've noticed a strange tingling sensation in my feet, and they've been feeling a bit swollen lately. I'm concerned about what this could be. | Foot Problems Flowchart | Numbness Or Tingling Flowchart | The user mentions tingling in their feet, which can be classified as a foot problem. However, the agent overemphasizes the user's strange tingling sensation, which leads to a misidentification that is more relevant to this symptom. | ☑ |
| | | | | Detailed | Female, 12 years | I'm a young student, and I've been having this strange, uncomfortable feeling in my chest. It's not exactly pain, but it's definitely not normal. It's been happening for a few days now, and it's been making me feel a bit anxious. I don't want to worry my parents, but I think it's important to get it checked out to make sure everything is okay. | Chest Pain Flowchart | Anxiety Flowchart | The user mentions discomfort in their chest, but they also feel anxious. Because of this, the agent prioritizes the patient's anxiety and misidentifies. | ☑ |
| Gemini Lite | Profile-Influenced | Because of the patient's profile, the system picked a more age-specific or gender-specific flowchart. | 1 (46.3%) | Brief | Female, 81 years | I've been feeling generally unwell and a bit confused lately. | Feeling Generally Ill Flowchart | Confusion In Older People Flowchart | The user mentions feeling generally unwell and confused. Because of the specific age provided, the agent identifies this as confusion in an older person as the final answer. | ☑ |
| | | | | Detailed | Male, 7 months | My baby boy has been acting a little strange lately. My wife and I have noticed that he doesn't seem to be following objects with his eyes as well as he used to. He's also been rubbing his eyes a lot. We're not sure if he's just tired or if he has a problem with his vision. We've tried to get him to look at toys, and he seems to struggle. We're very concerned and would appreciate some guidance. | Disturbed Or Impaired Vision Flowchart | Crying In Infants Flowchart | The user mentions that their son has been rubbing his eyes a lot and is unsure if it's related to a vision problem. However, because the patient is an infant, the agent prioritizes the discomfort of the infant. | ☑ |
| | Symptom Overlap | The opening statement generated for one flowchart contains similar keywords or symptoms that correlate with another flowchart. | 2 (36.5%) | Brief | Male, 15 years | I'm wheezing and my chest feels tight. I can't catch my breath. | Wheezing Flowchart | Difficulty Breathing Flowchart | The user mentions they are wheezing and that their chest feels tight. Both of these symptoms are related to wheezing and difficulty breathing, which leads the agent to identifying "Difficulty Breathing Flowchart" as the final flowchart. | ☑ |
| | | | | Detailed | Female, 32 years | I'm experiencing some pretty unpleasant symptoms. For the last few days, my stools have been really loose and watery, and I've had a lot of stomach cramps. I've also been feeling really tired and weak, and I've had to run to the bathroom several times a day. I haven't eaten anything unusual, and I'm not sure what could be causing this. I'm worried I might have some kind of infection or something. | Abnormal Looking Stools Flowchart | Diarrhea Flowchart | The user mentions unusual stools that are loose and watery, which can also be categorized as diarrhea, so the agent identifies "Diarheara Flowchart" as the final flowchart. | ☑ |
| | Confusing Opening | The opening includes details that distract from the main symptom or condition. As a result, the model may misidentify the user's intent, linking the case to another flowchart or returning "no flowchart available." | 3 (17.2%) | Brief | Male, 75 years | I've been coughing a lot after I eat. I think I'm aspirating my food. | Difficulty Swallowing Flowchart | no flowchart available | The user mentions difficulty with eating, specifically coughing and aspirating the food. Because of these specifics, the agent doesn't classify this as just difficulty in swallowing food, so it identifies no relevant flowchart. | ☑ |
| | | | | Detailed | Female, 18 years | I've been feeling really stressed lately with school and everything. I'm finding it hard to remember things I've studied, even right after I've read them. I'm also forgetting what I'm supposed to do for my homework. I'm not sure if it's just stress or something else, but I'm really worried that I'm going to fail my classes. I'm also having trouble sleeping, and I feel like I'm constantly distracted. I'm hoping you can help me figure out what's going on. | Impaired Memory Flowchart | Anxiety Flowchart | The user mentions having trouble remembering things. However, the agent also paid attention to the user being stressed, so it identifies this as an anxiety symptom instead of just impaired memory. | ☑ |
| k Chat | Profile-Influenced | Because of the patient's profile, the system picked a more age-specific or gender-specific flowchart. | 1 (44.1%) | Brief | Female, 6 months | Baby crying nonstop, pulling legs up. Seems like belly pain but also fussy when touched near chest. | Chest Pain Flowchart | Crying In Infants Flowchart | The user mentions their baby is fussy when their chest is touched, which may relate to chest pain. However, because their baby is an infant who is fussy, the agent identifies this as crying in an infant. | ☐ |
| | | | | Detailed | Female, 8 months | My baby has been extremely fussy for the last two days, constantly rubbing her left ear and crying during feedings. She's had a runny nose, and now she's warm; I checked her temperature, and it's slightly elevated. She's not sleeping well and seems uncomfortable whenever she moves her head. | Earache Flowchart | Crying In Infants Flowchart | The user mentions their baby being fussy and rubbing her ear, which are related to earache. However, because of the baby's age, the the agent prioritizes the baby's crying and chooses "Crying In Infants Flowchart". | ☑ |
| | Symptom Overlap | The opening statement generated for one flowchart contains similar keywords | 2 (39.7%) | Brief | Male, 72 years | I've been hearing whispers when no one's around. It's never happened before, and I'm freaked out. | Unusual Behavior Flowchart | Hallucinations Flowchart | The user mentions hearing whispers, which had never happened before (unusual). However, these symptoms are also related to hallucinations, so the agent identifies "Hallucinations Flowchart" as the final flowchart. | ☐ |

| | | | | | | | | | |
|---|---|---|---|---|---|---|---|---|---|
| Deepseek | Symptom Overlap | or symptoms that correlate with another flowchart. | 2 (39.7%) | Detailed | Male, 45 years | I've noticed a dark, irregularly shaped mole on my back that's grown over the past few months. It's not painful, but the edges look uneven, and I'm worried it might be something serious. My family has a history of skin cancer, so I wanted to get it checked. | General Skin Problems Flowchart | Raised Spots And Lumps Flowchart | The user mentions an unusal mole, which can be classified as a skin problem. However, moles can also be classified as spots and lumps, so the agent identifies "Raised Spots and Lumps Flowchart" as the final flowchart. | ☑ |
| | Confusing Opening | The opening includes details that distract from the main symptom or condition. As a result, the model may misidentify the user's intent, linking the case to another flowchart or returning "no flowchart available." | 3 (16.2%) | Brief | Female, 19 years | I've been feeling weird—dizzy and tired, but no obvious illness. Just not normal. | Feeling Generally Ill Flowchart | Dizziness Flowchart | The user mentions feeling weird, which related with feeling ill. However, because the user elaborates by saying they feel dizzy and tired, the agent places more emphasis on these terms and finds "Dizziness Flowchart" to be more fitting. | ☑ |
| | | | | Detailed | Male, 45 years | I've had a fever of 101°F for four days now, along with a persistent dry cough and shortness of breath. My chest feels heavy, and I'm exhausted. I don't smoke, but I'm worried this could be pneumonia. Should I get a chest X-ray, or is this viral? | Fever Flowchart | Difficulty Breathing Flowchart | The user mentions having a high fever. However, the agent places more emphasis on the user's shortness of breath and heavy chest, so "Difficulty Breathing Flowchart" is chosen as the final flowchart. | ☑ |

**Table 2: Generated demographic details and opening statement samples from four language models**

| Model | Flowchart | Flowchart Description | Pattern | Sex | Age | Opening Statement |
|---|---|---|---|---|---|---|
| OpenAI GPT4o | Infertility In Men Flowchart | All ages - Male - Flowchart for the inability to conceive after more than 12 months of having intercourse without using contraception. | Brief | Male | 50 years | Trying for a child for 15 months without success. Do I need to worry about infertility? |
| | Abnormal Looking Stools Flowchart | All ages - Male and Female - Flowchart for stools that are not the usual color or consistency. | Brief | Female | 19 years | I've noticed my stools are a weird texture. Is there something wrong with my diet? |
| | Nightmares Flowchart | All ages - Male and Female - Flowchart for frightening dreams that may be disturbing enough to wake you. | Brief | Male | 12 years | My son has nightmares that wake him up scared. How can we help him? |
| | Headache Flowchart | All ages - Male and Female - Flowchart for mild to severe pain in the head. | Brief | Female | 7 years | My daughter has been complaining of a mild headache all day. Should I be concerned? |
| | Feeling Generally Ill Flowchart | All ages - Male and Female - Flowchart for a vague sense of not being well. | Brief | Male | 77 years | Lately, I've felt a bit off, like something's not quite right. |
| | Unusual Behavior Flowchart | All ages - Male and Female - Flowchart for behavior that is significantly different from a person's usual behavior. | Detailed | Male | 5 years | Hello, I'm concerned about my son. He's been acting differently, like throwing tantrums over small things and being more clingy than usual. This isn't how he normally behaves, and I'm worried there might be an underlying issue. What should I do to help him? |
| | Disturbed Or Impaired Vision Flowchart | All ages - Male and Female - Flowchart for vision problems, including blurring, double vision, or seeing flashing lights or floating spots. | Detailed | Female | 45 years | Hi, I'm a 45-year-old woman and recently, I've noticed that my vision has become impaired. I often see flashing lights in my peripheral vision, which is quite alarming. This has been happening for a few days now, and it's accompanied by occasional headaches. I'm worried about what this could mean for my eye health. |
| | Hearing Loss Flowchart | All ages - Male and Female - Flowchart for impaired ability to hear in one or both ears. | Detailed | Male | 32 years | I've been having trouble hearing out of my left ear for the past few weeks. It's like there's a constant ringing, and I'm finding it hard to follow conversations, especially in noisy environments. I've also noticed some dizziness and occasional headaches. Could this be something serious? |
| | Waking At Night In Children Flowchart | 0-5 years - Male and Female - Flowchart for difficulty sleeping through the night that causes a child to cry or call out. | Detailed | Female | 18 months | Good day, my 18-month-old daughter has started waking up at night, crying and calling out. She seems very distressed and won't go back to sleep easily. This started about two weeks ago, and we're at a loss for what might be causing it. Could this be separation anxiety or something else? |
| | Rash With Fever Flowchart | All ages - Male and Female - Flowchart for spots, discolored areas, or blisters on the skin and a temperature of 100f or higher. | Detailed | Male | 18 years | Hey, I'm a bit worried because I've got these small blisters on my arms and a fever of 102°F. The blisters are filled with clear fluid and are quite painful. I've also been feeling really tired and have lost my appetite. Could this be a sign of something infectious? |
| ku | Painful Or Enlarged Testicles Flowchart | All ages - Male - Flowchart for pain or swelling that may affect one or both testicles or the scrotum (the pouch that contains the testicles). | Brief | Male | 39 years | I've been experiencing a heavy, aching sensation in my scrotum, and it's been bothering me for the past few days. What could this mean? |
| | Feeling Generally Ill Flowchart | All ages - Male and Female - Flowchart for a vague sense of not being well. | Brief | Female | 54 years | I've been experiencing a general sense of unease, and I'm not sure what's causing it. Can you help me figure out the next steps? |
| | Chest Pain Flowchart | All ages - Male and Female - Flowchart for any pain between the neck and the bottom of the rib cage. | Brief | Female | 8 years | I told my mom that my tummy hurts, and she's worried it might be something serious. Can you help me figure out what's wrong? |
| | Raised Spots And Lumps Flowchart | All ages - Male and Female - Flowchart for raised areas on the skin that may be inflamed, dark-colored, rough, or hard. | Brief | Male | 9 months | My baby has developed some raised, discolored areas on their skin. I'm worried it could be a skin condition. Can you advise me? |
| | Difficulty Speaking Flowchart | All ages - Male and Female - Flowchart for difficulty choosing, using, or pronouncing words. | Brief | Female | 83 years | I'm finding it increasingly challenging to use the right words when I'm trying to express myself, and it's really frustrating. |
| | Difficulty Sleeping Flowchart | >5 years - Male and Female - Flowchart for frequent problems falling asleep or staying asleep. | Detailed | Male | 54 years | I'm a 54-year-old man, and I've been having a lot of trouble sleeping lately. It takes me forever to fall asleep, and then I wake up multiple times throughout the night, often feeling restless and unable to go back to sleep. This is really impacting my daily life, as I'm constantly feeling tired and have a hard time focusing at work and with my hobbies. I'm wondering if there are any lifestyle changes or medical interventions that could help me improve my sleep quality. |
| Claude Hai | Painful Leg Flowchart | All ages - Male and Female - Flowchart for intermittent or continuous pain in the thigh or calf. | Detailed | Female | 6 months | My baby daughter has been crying inconsolably and seems to be in pain in her right leg. She's been refusing to put any weight on it, and it appears to be causing her a lot of discomfort. We're very concerned and want to get her checked out by a pediatrician as soon as possible to make sure there's nothing seriously wrong and to get her the appropriate treatment. |
| | Recurring Abdominal Pain Flowchart | >12 years - Male and Female - Flowchart for abdominal pain that comes and goes. | Detailed | Female | 22 years | I'm a young, otherwise healthy individual, but for the past several months, I've been experiencing these recurring episodes of abdominal pain. The pain can be quite intense, and it often comes on without any warning. It's usually centered in the middle of my abdomen, and it can radiate to my back and sides. I've also noticed that the pain is sometimes accompanied by bloating, gas, and changes in my bowel movements. I'm really worried that there might be something more serious going on, and I'd like to get it checked out as soon as possible. |
| | Painful Eye Flowchart | All ages - Male and Female - Flowchart for continuous or intermittent pain in or around the eye. | Detailed | Male | 3 years | My 3-year-old son has been crying and rubbing his right eye a lot lately. He seems to be in pain, and the eye also appears to be red and swollen. He's not old enough to tell me exactly what's wrong, but I'm concerned that he might be experiencing some kind of eye issue. I'd like to get him checked out by a doctor to make sure there's nothing serious going on. |
| | Hallucinations Flowchart | All ages - Male and Female - Flowchart for abnormal sensory perceptions that occur without an external stimulus and are not based on reality. | Detailed | Male | 55 years | I'm a business executive, and for the past few months, I've been experiencing some really bizarre sensory perceptions. I'll be in a meeting, and suddenly I'll smell something that isn't there, like the scent of a campfire or freshly cut grass. And sometimes I'll hear sounds that no one else can hear, like a faint tapping or a whisper. It's really starting to affect my work and my daily life, and I'm worried that it might be a sign of a neurological condition or a mental health issue. Can you help me figure out what's going on? |
| Gemini Lite | Itchy Spots And Rashes Flowchart | >2 years - Male and Female - Flowchart for discolored or raised areas of itchy skin. | Brief | Female | 12 years | I have these itchy spots that are spreading. They're really bothering me, and I don't know what to do. |
| | Abdominal Pain Flowchart | >12 years - Male and Female - Flowchart for pain between the bottom of the rib cage and the groin. | Brief | Male | 78 years | I have a sudden and intense pain in my abdomen. I'm very concerned. |
| | Painful Knee Flowchart | All ages - Male and Female - Flowchart for pain in or around the knee joint. | Brief | Male | 35 years | I hurt my knee playing basketball yesterday. It's hard to put weight on it. |
| | Runny Nose Flowchart | All ages - Male and Female - Flowchart for a partially or completely blocked nose, with a liquid discharge. | Brief | Female | 1 month | My baby has a runny nose and seems a little congested. Should I be worried? |
| | Coughing In Children Flowchart | 2-12 years - Male and Female - Flowchart for coughing in children, which is usually a symptom of a respiratory infection. | Brief | Male | 11 years | My son's cough is getting worse. He's also having trouble breathing sometimes. |
| | Facial Skin Problems Flowchart | >2 years - Male and Female - Flowchart for any rash, spots, or changes in the skin on the face. | Detailed | Female | 28 years | I'm experiencing a sudden onset of redness and swelling on my face, particularly around my eyes and mouth. It started a few hours ago and seems to be getting worse. It's also quite itchy, and I'm starting to feel a little lightheaded. I haven't used any new skincare products or eaten anything unusual that I'm aware of. I'm a little worried it could be an allergic reaction, and I'm not sure what to do. I also have a slight tingling sensation in my lips. |
| | Pain In The Face Flowchart | All ages - Male and Female - Flowchart for pain in one or both sides of the face or forehead. | Detailed | Female | 45 years | I've been dealing with a sharp, shooting pain on the left side of my face, specifically around my cheek and jawline. It comes and goes, triggered by things like chewing, talking, and even a slight breeze. It started a few weeks ago, and I've been trying to ignore it, but it's getting more frequent and intense. I don't have any other symptoms, like fever or swelling, but the pain is really bothersome, and I'm starting to get worried about it. |
| | Feeling Faint And Fainting Flowchart | All ages - Male and Female - Flowchart for a sudden feeling of weakness and unsteadiness that may result in brief loss of consciousness. | Detailed | Male | 8 years | I was playing video games, and I got up to get a snack. Everything went black, and I fell on the floor! My mom said I was out for a few seconds. My head hurts a little bit now, and I feel a little bit sick. I'm not sure what happened. I had a big breakfast this morning, but it's been a long time since I've eaten anything. I'm scared to play again in case it happens again. |
| | Hoarseness Or Loss Of Voice Flowchart | All ages - Male and Female - Flowchart for abnormal huskiness in the voice. | Detailed | Male | 35 years | Good morning, I'm a bit concerned about my voice. I'm a teacher, and I've been noticing my voice is hoarse and strained, especially at the end of the school day. It feels like I'm constantly clearing my throat, and sometimes it feels like I'm losing my voice altogether. This has been going on for about a week now, and it's affecting my ability to do my job. I've been trying to drink more water and rest my voice, but it doesn't seem to be helping. I don't have a cold or any other symptoms, just this persistent hoarseness, and it's making it hard to get through my lessons. |
| | Skin Problems In Young Children Flowchart | 0-2 years - Male and Female - Flowchart for skin discoloration, inflammation, or blemishes. | Detailed | Male | 6 months | Hi, I'm really worried about my baby boy. He's only six months old, and I've noticed these red, bumpy patches appearing on his cheeks and around his mouth. It seems to be getting worse, especially after he eats. He doesn't seem to be in any pain, but he's definitely scratching at them sometimes, and I'm afraid it might be bothering him. We've tried a bit of baby lotion, but it doesn't seem to be helping. Is this something I should be concerned about, or is it just a normal rash that will go away on its own? |
| Deepseek Chat | Painful Or Enlarged Testicles Flowchart | All ages - Male - Flowchart for pain or swelling that may affect one or both testicles or the scrotum (the pouch that contains the testicles). | Brief | Male | 60 years | My scrotum is enlarged and tender. No fever, but discomfort for days. |
| | Bad Breath Flowchart | All ages - Male and Female - Flowchart for foul-smelling breath that may be temporary or persistent. | Brief | Female | 28 years | I've had bad breath for a week, and brushing isn't helping. Could it be something serious? |
| | Wheezing Flowchart | All ages - Male and Female - Flowchart for noisy, difficult breathing. | Brief | Female | 6 months | My baby is making whistling sounds when breathing and seems fussy. Should I be worried? |
| | Vaginal Irritation Flowchart | All ages - Female - Flowchart for itching or soreness inside or just outside the vagina. | Brief | Female | 22 years | I have a weird itch inside my vagina and a strange smell. What could this be? |
| | Earache Flowchart | All ages - Male and Female - Flowchart for pain in one or both ears. | Brief | Male | 28 years | I've had sharp pain in my left ear for two days, and it's getting worse. |
| | Itchy Spots And Rashes Flowchart | >2 years - Male and Female - Flowchart for discolored or raised areas of itchy skin. | Detailed | Female | 4 years | My daughter has been scratching these rough, scaly patches on her elbows and knees for over a month. They're pinkish and seem to flare up after baths. She's uncomfortable and can't sleep well. We've tried moisturizers, but nothing helps. Could this be eczema or an allergic reaction? |
| | Diarrhea Flowchart | >6 months - Male and Female - Flowchart for frequent passing of unusually loose stools. | Detailed | Female | 28 years | Since last summer, I've been having loose stools almost daily, often with mucus. I've tried cutting out dairy and gluten, but it hasn't helped. I also get sharp abdominal pains and feel gassy all the time. It's embarrassing and stressful-I'd like to know if this could be IBD or something else. |
| | Numbness Or Tingling Flowchart | All ages - Male and Female - Flowchart for loss of feeling or a prickly sensation in any part of the body. | Detailed | Male | 68 years | Over the last few months, I've had this persistent numbness in both feet-it's like walking on cotton wool. I have diabetes, so I'm worried it might be neuropathy. Sometimes, I get sharp, burning pains too. Should I see someone soon, or is this something I can manage at home? |
| | Abnormal Looking Urine Flowchart | All ages - Male and Female - Flowchart for urine that differs from its usual color or that is cloudy or tinged with blood. | Detailed | Male | 45 years | Good day, I'm a bit concerned about my urine. I noticed a few blood clots in it this morning, and it was a bright red color. I also have a sharp pain in my side, just below my ribs, that comes and goes. I've been feeling a bit nauseous, and I had a bit of a fever last night. I've never experienced anything like this before, and I'm a bit worried about kidney stones or something similar. I've been trying to drink more water, but the symptoms haven't improved. |
| | Hair Loss Flowchart | All ages - Male and Female - Flowchart for thinning of hair or hair loss on all or part of the head. | Detailed | Female | 28 years | After having my baby six months ago, my hair has been shedding like crazy. It's everywhere-on my pillow, in the drain, even in my baby's hands. I know postpartum hair loss is common, but this seems excessive. Could it be related to my thyroid? I've also been feeling unusually tired. |

**Table 3: Failure analysis for flowchart navigation evaluation task**

| Pattern | Reasons of Failure | Occurrence | Failure Examples | | | | | | | | |
|---|---|---|---|---|---|---|---|---|---|---|---|
| | | | Flowchart | Question from Flowchart | User Response | Answer Label | Decision Agent Output | | | | Explanation of Failure |
| | | | | | | | isOnTopic | isAnswered | isUncertain | actualAnswer | |
| 1: Brief | User response contains a misleading or unclear way of agreeing, which results in no definitive answer. | 68.0% | Foot Problems Flowchart | Are you over age 50 and do you also have pain in your ankle, knee, or hip? | Absolutely. | Yes | Yes | No | No | N/A | Because the response pattern is defined as brief, it limits the length of the response.The question contains two parts connected by "and", but the response provides only a single word due to the predefined pattern type. As a result, the ambiguity in agreement leads the agent to categorize it as "unanswered". |
| | | | Abnormal Looking Urine Flowchart | Is your urine dark yellow or orange? | Uh-huh. | Yes | Yes | No | No | N/A | The informal acknowledgment “uh-huh” does not clearly convey affirmation or negation, leading the agent to interpret the response as ambiguous and categorize it as “unanswered.” |
| | The question asked to the patient is an inclusive OR, but the agent interprets it as an exclusive OR, which leads to a wrong interpretation. | 16.0% | Nightmares Flowchart | Have you recently been drinking more alcohol than usual, or have you recently stopped drinking? | That's right. | Yes | No | No | No | N/A | The question contains two parts connected by “or,” but the response provides only a single word due to the predefined pattern type. It is unclear whether the response is affirming both parts the response. Additionally, it doesn't include any information related to the question due to the briefness, so the agent categorizes it as "off-topic". |
| | | | Chest Pain Flowchart | Have you recently had chest surgery or a chest injury, or have you been coughing severely? | Yeah, a bad cough for days now. | Yes | Yes | Yes | No | No | The question contains multiple parts connected by “or,” but the response only addresses one (“coughing severely”) due to the predefined pattern type. And because the agent interprets the "or" exclusively, it incorrectly classifies the response as "No". |
| 2: Descriptive | The question asked to the patient is an inclusive OR, but the agent interprets it as an exclusive OR, which leads to a wrong conclusion. | 58.6% | Painful Ankles Flowchart | Are both ankles or any other joints (such as your knee or finger joints) affected? | Absolutely, my knees have been giving me a lot of trouble lately. The joints feel swollen and achy, especially after I've been on my feet for a while. | Yes | Yes | Yes | No | No | The response clearly affirms joint are affected, but doesn't mention ankles. Because the agent interprets the "or" exclusively, it misclassifies the response as "No". |
| | | | Abnormal Looking Stools Flowchart | Have you been vomiting or have you had diarrhea? | Yeah, I've been dealing with diarrhea for the past two days. No vomiting, but my stomach is really upset. | Yes | Yes | Yes | No | No | The response clearly affirms diarrhea, but denies vomiting, and because the agent interprets the "or" in the question exclusively, it incorrectly classifies the response as "No". |
| | The patient response contains confusing details. | 23.0% | Difficulty Sleeping Flowchart | Do you get little or no exercise on most days? | No, I don't get much exercise. I used to go for walks regularly, but lately, I've been so busy that it's fallen by the wayside. | Yes | Yes | Yes | No | No | The response affirms the question overall but uses a negative construction "No, I don't get much exercise". The agent interprets the initial "No" literally as a negation, leading it to misclassify the response as "No" instead of "Yes". |
| | | | Bad Breath Flowchart | Do you brush and floss your teeth less often than twice a day? | Honestly, no. I brush once a day, usually in the morning, and I try to floss every other day, but it's not a daily habit. | Yes | Yes | Yes | No | No | The response affirms the question overall but begins with a negative cue "No", which confuses the agent. The agent interprets the initial "No" as disagreement, even though the details that follow ("brush once a day", "floss every other day") indicate a "Yes". |
| 3: Weak | The agent overestimates the user’s certainty and ignores ambiguity because it assumes any affirmative or negative cue reflects a confident, final intent. | 67.7% | Constipation Flowchart | Are bowel movements painful? | I wouldn't exactly call them pleasant, if you know what I mean. There's definitely some discomfort there. | Yes | Yes | Yes | No | No | The response affirms the presence of pain through implied language ("wouldn't call them pleasant"), but the agent fails to detect this subtle uncertainty. And because the statement lacks a definite affirmation, it is interpreted as a confident "No". |
| | | | Swellings Under the Skin Flowchart | Is the lump or swelling painful, red, and warm? | Kind of...it's more achy than sharp, and the area seems a little flushed. | Yes | Yes | Yes | No | No | The response expresses a weak "Yes" through hedged language ("Kind of...") and additional description about the symptoms. However, the agent fails to recognize the uncertainty cue and the implicit affirmation ("achy", "a little flushed"), interpreting the response as a definite "No" instead of an uncertain "Yes". |
| 4: Uncertain | The generated patient response indicates some level of affirmation or denial, so the agent identified that an answer is present. | 68.6% | Cramp Flowchart | Did the cramp occur while you were sleeping? | I can't really pinpoint when it started. I was asleep, and then I just found myself awake with the cramp. | not answered | Yes | Yes | No | Yes | The response conveys uncertainty about the timing of the cramp, indicating that the patient cannot confirm or deny whether it began during sleep. However, the agent overlooks the uncertainty cue and interprets the contextual mention of "being asleep" and "awake with the cramp" as a definite "Yes". |
| | | | Slow Weight Gain in Young Children Flowchart | Is your baby on a rigid breastfeeding schedule? | I wouldn't say it's rigid; we kind of go with the flow and adjust as needed. | not answered | Yes | Yes | No | No | The response conveys uncertainty and partial disagreement. "I wouldn't say it's rigid" softens the statement and implies flexibility rather than complete negation. The agent interprets it as a confident "No", overlooking the noncommittal tone. |
| 5: Off-topic | The patient response begins by addressing the question, though it later introduces some irrelevant details. It is still considered on-topic rather than off-topic. | 97.0% | Swellings Under the Skin Flowchart | Are you currently taking any medication? | Medication? Goodness, no! I much prefer to let my inner child out to play. Have you ever built a pillow fort? It's quite therapeutic, I must say. | off-topic | Yes | Yes | No | No | The response is supposed to be completely off-topic, while it first answers the question with a "No", and then introduces irrelevant information. The agent classifies it as a definite No due to the first part of the response. |
| | | | Sore Mouth or Tongue Flowchart | Did you recently start using any new cosmetics or lotions on your lips? | No, but the weather has been so strange lately, hasn't it? One minute it's sunny, the next it's pouring. | off-topic | Yes | Yes | No | No | The response is supposed to be completely off-topic, while it first answers the question with a "No", and then introduces irrelevant information. The agent classifies it as a definite No due to the first part of the response. |

**Table 4: Generated patient response samples from four language models**

| Model | Pattern | Flowchart | Question from Flowchart | Answer Label | Generated Patient Response |
|---|---|---|---|---|---|
| OpenAI GPT4o | 1: Brief | Abnormal Hair Growth In Women Flowchart | Is the excessive hair mainly on your face? | Yes | Yep. |
| | | Crying In Infants Flowchart | Does your baby stop crying after being fed? | Yes | Oh yes. |
| | | Hearing Loss Flowchart | Does your ear have a sticky yellow discharge? | No | Nope. |
| | | Limping In Children Flowchart | Does your child have a fever and a painful, tender area over any bone in the leg or foot? | Yes | For sure, yes. |
| | | Painful Or Enlarged Testicles Flowchart | Have you suddenly developed a painful swelling in one or both testicles? | No | Not at all. |
| | 2: Descriptive | Abnormal Looking Urine Flowchart | Have you eaten any foods within the past 24 hours that contain red or dark artificial dyes (such as brightly colored candies) or that are dark in color (such as rhubarb, beets, or blackberries)? | Yes | Definitely, I had a bowl of mixed berries, including blackberries, for breakfast, and I also ate some red Skittles in the afternoon. |
| | | Difficulty Breathing Flowchart | Is your temperature 100°F or above, or are you coughing up greenish yellow or rust-colored phlegm? | Yes | Absolutely, I've been running a fever of 100.5°F all day, and it's making me feel pretty lousy. On top of that, my cough has gotten worse, and the phlegm is definitely a rusty color. |
| | | Impaired Memory Flowchart | Are you currently taking any medication? | No | No, I'm not taking any medication at the moment. I try to manage everything with diet and exercise, and so far, it's been working well for me. |
| | | Noises In The Ear Flowchart | Do you have a tickling sensation in your ear? | Yes | Yes, there's a tickling in my ear, almost like something's moving around in there. It's not painful, but it's really distracting. |
| | | Painful Periods Flowchart | Does the pain get worse as your period continues? | Yes | Yes, it does. I notice that by the middle of my period, the cramps are so bad that I have to take time off work because I can't concentrate on anything else. |
| | 3: Weak | Abnormally Frequent Urination Flowchart | Have you noticed two or more of the following symptoms: increased thirst; weight loss; unexplained tiredness; blurred vision? | Yes | I think so, yeah. I've been feeling more tired than usual and maybe a bit thirstier, but I'm not entirely sure if it's just because I've been busy lately. |
| | | Dizziness Flowchart | Did you injure your head recently? | Yes | Yeah, I suppose you could say that. I had a bit of a mishap and ended up hitting my head, though I'm not sure how much of an injury it really is. |
| | | Irregular Vaginal Bleeding Flowchart | Was the bleeding like that of a period? | Yes | Yeah, it kind of felt like a period, but there was something a bit off about it. |
| | | Pain In The Face Flowchart | Do you have stabbing pain on one side of your face, brought on by any of the following: touching your face; chewing; breathing cold air; drinking cold liquid? | No | Not really, at least not that I've paid attention to, but I guess I could have missed something. |
| | | Palpitations Flowchart | Do you feel ill, do you have any chest discomfort, or have you had irregular heartbeats or other heart problems? | No | I haven't noticed anything major, but I can't say for sure if everything's perfect. |
| | 4: Uncertain | Bad Breath Flowchart | Do you brush and floss your teeth less often than twice a day? | not answered | I aim for twice a day, but life happens, and I might not always stick to that. It's kind of hit or miss. |
| | | Fever Flowchart | Do you have a headache, or do your bones and joints ache? | not answered | I'm feeling a bit under the weather, but I can't figure out if it's my head or my bones that are bothering me. |
| | | Itching In Children Flowchart | Is the itching mainly in an area covered by clothing? | not answered | It's hard to say, sometimes it feels like it might be, but other times it seems more spread out. |
| | | Painful Ankles Flowchart | Are you over age 50? | not answered | Well, let's just say I've been around the block a few times, but I like to keep people guessing about my exact age. |
| | | Raised Spots And Lumps Flowchart | Do you have a slow-growing, dark-colored lump, or have you noticed a change in a mole? | not answered | I'm not really sure, to be honest. I haven't paid much attention to any changes, but I guess I could take a closer look to see if anything seems off. |
| | | Confusion Flowchart | Were you drinking alcohol or taking other drugs before you became confused? | off-topic | I was actually just trying to remember the name of that actor from that movie we watched last week. It's on the tip of my tongue! |
| Claude Haiku | 5: Off-topic | Foot Problems Flowchart | Is there an area of redness or swelling on the sole of your foot, heel pain, or a swollen area at the base of your big toe? | off-topic | Did you see the latest episode of that TV show everyone's talking about? It had such a surprising twist! |
| | | Lack Of Bladder Control Flowchart | Do you have pain when you urinate? | off-topic | My cat has been acting so funny lately, chasing her tail all around the house. |
| | | Painful Intercourse In Men Flowchart | Do you have any redness, swelling, lumps, or sores on the skin or tip of your penis, or does your penis bend at an angle during an erection? | off-topic | I've been really into cooking lately, trying out new recipes and experimenting with different spices. |
| | | Recurring Vomiting Flowchart | Do the whites of your eyes and your skin look yellow? | off-topic | You know, I've been thinking about trying out that new Italian restaurant downtown. Have you been there yet? |
| | 1: Brief | Abnormal Hair Growth In Women Flowchart | Do you have two or more of the following symptoms: unexplained weight gain; deepening voice; absence of periods? | No | No, none of those apply to me. |
| | | Coughing In Children Flowchart | Does anyone in the house smoke, or could your child be smoking? | Yes | Yes, someone in the house smokes. |
| | | Impaired Memory Flowchart | Are you currently taking any medication? | Yes | Yes, I am. |
| | | Painful Arm Or Hand Flowchart | Do you have stiffness in the morning that lasts longer than an hour? | No | No, I don't have that. |
| | | Swollen Abdomen Flowchart | Could you be pregnant? | Yes | Yes, there's a chance I could be. |
| | 2: Descriptive | Abnormal Looking Urine Flowchart | Is your urine dark yellow or orange? | Yes | Yes, that's correct. My urine has taken on a more golden, almost tangerine hue recently. It's noticeably darker and more vibrant than it typically is. |
| | | Diarrhea In Infants Flowchart | Is your baby's temperature over 100°F, and is he or she vomiting? | No | No, neither high temperature nor vomiting is an issue for my baby right now. |
| | | Infertility In Women Flowchart | Is either of you ill, or does either of you have a chronic disease? | No | No, that's not the case. We're both feeling well and don't have any chronic diseases that we're aware of. |
| | | Painful Intercourse In Women Flowchart | Do you have persistent itching around the genital area? | No | Thankfully, no. I haven't experienced any persistent or bothersome itching around my genitals. Everything seems to be functioning properly down there. |
| | | Toothache Flowchart | Have you had one or more teeth filled within the past week? | Yes | Yes, I actually just had a filling done on one of my molars last Tuesday. The dentist said it was a small cavity that needed to be taken care of. |
| | 3: Weak | Abnormally Frequent Urination Flowchart | Do you sometimes have a strong urge to urinate followed by an uncontrollable leaking of urine? | No | No, I don't think that's something I've dealt with. It doesn't sound familiar to me, but I suppose it's possible it's happened and I just don't remember. I can't say for sure either way. |
| | | Difficulty Sleeping Flowchart | Do you wake up during the night or very early in the morning and find it difficult to fall asleep again? | Yes | I guess you could say that's a problem I deal with. I often find myself waking up in the middle of the night or really early in the morning, and then just lying there unable to fall back asleep. |
| | | Itching In Children Flowchart | Is the itching mainly in an area covered by clothing? | No | Well, it's hard to say for sure, but I don't think the itching is just in the covered areas. |
| | | Painful Or Stiff Neck Flowchart | Have you had a problem controlling an arm or leg since then? | Yes | I think I've been having a bit of trouble controlling my arm/leg, to be honest. It's been acting up on and off, but I can't say for certain if it's a real issue. |
| | | Twitching And Trembling Flowchart | Is the trembling or shaking limited to one part of your body? | Yes | It does seem to be focused in one spot, like my shoulder, but I can't say for certain that it's not affecting anything else as well. |
| | 4: Uncertain | Backache Flowchart | Does the pain shoot down the back of your leg? | not answered | You know, I'm not entirely sure about that. The pain is definitely present, but I can't quite pinpoint whether it's traveling down my leg or not. Let me think about that a bit more. |
| | | Disturbed Or Impaired Vision Flowchart | Is only one eye affected? | not answered | Hmm, that's a good question. To be honest, I'm not entirely sure. The issue seems to be centered in one eye, but I can't say for certain if the other one is unaffected. I'd need to get a better look to know for sure. |
| | | Lack Of Bladder Control Flowchart | Do you have pain when you urinate? | not answered | Well, it's a little hard to tell. I've been feeling a bit of a burning sensation, but I'm not sure if I'd call it outright pain. It's a bit of an uncomfortable feeling, but I can't say for sure if it's painful or not. |
| | | Painful Urination Flowchart | Are you urinating more frequently than usual? | not answered | I'm not entirely sure. It's possible I've been going a bit more often, but it's hard for me to say for sure without really keeping better track of it. |
| | | Unusual Behavior Flowchart | Could the person be abusing alcohol or other drugs? | not answered | Hmm, it's possible, but I can't really tell from what I know so far. There could be other explanations too. I'd need to get a better sense of the full situation to make a call on that. |
| | 5: Off-topic | Confusion Flowchart | Have you had any of the following symptoms since the confusion started: dizziness; weakness in your arms or legs; numbness or tingling in any part of your body; blurred vision; difficulty speaking; abnormal movements of the body or face? | off-topic | Oh my, that's quite a list of potential issues. But you know, I've been so preoccupied with planning my upcoming vacation, I'm not sure I've been paying as much attention to my health as I should. Where do you think I should go, if you don't mind me asking? |
| | | Facial Skin Problems Flowchart | Do you have a dark lump or patch on your face, or has a mole changed in any way? | off-topic | Hmm, that's an interesting question. Have you ever noticed how the clouds outside seem to be shaped like different animals? I find cloud-watching to be quite soothing. |
| | | Noises In The Ear Flowchart | Do you have any hearing loss? | off-topic | Hearing loss? Not that I'm aware of. Although, come to think of it, my sense of smell has been a bit off lately. Do you have any tips for improving my olfactory senses? |
| | | Raised Spots And Lumps Flowchart | Do you have several rough-surfaced, hard lumps on your hands or feet? | off-topic | Rough, hard lumps - that doesn't sound very pleasant. Say, do you have any plans for the weekend? I was thinking of going on a hike, if the weather cooperates. |
| | | Vaginal Irritation Flowchart | Does the skin around your vagina look abnormal? | off-topic | The skin around my vagina? Oh, I'm sorry, I thought you were asking about something else. Did you know that the average person blinks around 15-20 times per minute? Fascinating stuff, isn't it? |
| | 1: Brief | Abdominal Pain In Children Flowchart | Has your child not had a bowel movement in 2 or 3 days, or does he or she have difficulty passing stools? | Yes | Yep, that's the situation. |
| | | Confusion In Older People Flowchart | Is the confusion accompanied by signs of illness such as a fever, coughing, or lack of bladder control? | No | Not at all. |
| | | Disturbed Or Impaired Vision Flowchart | Do you have a headache? | No | No, I don't. |
| | | Hoarseness Or Loss Of Voice Flowchart | Has the hoarseness or loss of voice lasted longer than a week? | Yes | Yep. |
| | | Sore Mouth Or Tongue Flowchart | Is your tongue sore? | Yes | Yeah. |
| | 2: Descriptive | Abnormal Looking Stools Flowchart | Do the whites of your eyes and your skin look yellow? | No | Definitely not. My skin and eyes are clear. I haven't noticed any change in color. I'm feeling fine otherwise too, no other symptoms that would suggest it. |
| | | Coughing In Children Flowchart | Is your child's temperature over 100°F? | Yes | Unfortunately, yes. It's hovering around 100.8°F. We've been trying to bring it down with some Tylenol, but it hasn't really budged yet. |
| | | Earache Flowchart | Has your hearing worsened in the past few weeks or months? | Yes | Unfortunately, yes. I think so. I've had to turn up the volume on the TV a lot higher than usual lately, and my wife keeps telling me I'm missing parts of conversations. It's really noticeable in the last month or so. |
| | | Infertility In Women Flowchart | Have you ever had an infection of the uterus or fallopian tubes or a sexually transmitted disease? | No | Nope, nothing like that. I've never had any infections of that nature. I've always practiced safe sex and been diligent about my health. |
| | | Swollen Ankles Flowchart | Is the calf of the same leg swollen, or do you feel pain in the calf when you walk? | No | Nah, no swelling or pain there. I've been walking around all morning, and everything feels normal in my calf. The issue I'm having is more in my ankle, actually. |
| | | Absent Periods Flowchart | Have you recently been ill or under stress? | No | Not really, I think I've been alright. Nothing major, at least. |
| ini Lite | | Crying In Infants Flowchart | Could your baby be cold? | Yes | Could be, yeah. She's not *super* warm to the touch, but it's hard to say for sure. |
| Gemi | 3: Weak | Facial Skin Problems Flowchart | Do you have a blistering rash on one side of your face in an area that was painful over the past 2 to 4 days? | Yes | Well, it *looks like* there's something going on on one side, and yeah, it's been a bit tender the last few days. I think so, yes. |
| | | Itching Without A Rash Flowchart | Have you recently had severe diarrhea? | No | I don't think so. There might have been a slight change in my bowel habits, but it hasn't been diarrhea, exactly. |
| | | Unusual Behavior Flowchart | Does the person seem preoccupied with a single idea or activity? | No | Hmm, I'm not sure. They haven't been fixated on anything in particular, as far as I can tell. |
| | 4: Uncertain | Anxiety Flowchart | Have you lost weight, or do your eyes seem to be bulging? | not answered | I haven't weighed myself recently, so I couldn't say. As for my eyes, I guess they might look a little different, but I'm not sure if it's bulging. |
| | | Diarrhea Flowchart | Have you felt ill or have you been vomiting? | not answered | I'm not sure. I feel a little off, but I haven't actually thrown up. |
| | | Gas And Belching Flowchart | Is the gas expelled by belching? | not answered | I guess so? I mean, it's definitely coming *out* somehow... but I don't know if that's the official cause. |
| | | Pain In The Face Flowchart | Do you have continuous, throbbing pain on one side of your face? | not answered | Well, it's definitely *something* on that side of my face. It comes and goes, and sometimes it feels like a dull ache, but... throbbing? I'm not sure I'd describe it that way exactly. |
| | | Waking At Night In Children Flowchart | Does your baby go back to sleep after the feeding? | not answered | It depends on the time of day, I think. During the day, maybe. At night, usually... but not always. |
| | 5: Off-topic | Breast Problems In New Mothers Flowchart | Can you see or feel one or more lumps in your breast? | off-topic | I'm really craving a good cup of tea right now. Earl Grey, maybe? |
| | | Difficulty Sleeping Flowchart | Did you eat a lot late in the day, or drink a lot of alcohol? | off-topic | This weather is something else, isn't it? A real change from yesterday. |
| | | Hair Loss Flowchart | Did the thinning occur 2 to 3 months after a fever? | off-topic | My cat has been acting so weird. He keeps staring at the wall and meowing at nothing. It's quite amusing, actually. |
| | | Painful Eye Flowchart | Does your eye feel gritty? | off-topic | I'm really craving a cup of coffee right now. |
| | | Wheezing Flowchart | Do you cough up gray or greenish yellow phlegm on most days? | off-topic | I'm trying to decide what to make for dinner tonight. Maybe something with pasta? |
| | 1: Brief | Abdominal Pain In Children Flowchart | Did your child seem healthy before the abdominal pain started? | Yes | Yes, totally fine. |
| | | Chest Pain Flowchart | Is the pain pressing or crushing, or does it radiate out from your chest to other parts of your body (such as your breastbone, the upper part of your abdomen, or your jaw, neck, or arms)? | Yes | Yeah, it spreads to my arm. |
| | | Difficulty Breathing Flowchart | Is the pain worse when you inhale? | No | Nope. |
| | | Gas And Belching Flowchart | Are your stools pale, greasy, and foul-smelling? | No | No, everything looks fine. |
| | | Overweight Flowchart | Are you over 40? | Yes | Yep. |
| | 2: Descriptive | Abnormal Looking Stools Flowchart | Do the whites of your eyes and your skin look yellow? | Yes | Yeah, I noticed it a few days ago. My eyes are definitely yellow, and my skin has a weird yellow tone, like a light tan but sickly. |
| | | Constipation Flowchart | Have you used laxatives regularly for a long time? | No | No, I've only used them once or twice in the past when really needed, but never long-term or frequently. |
| | | Difficulty Speaking Flowchart | Have you had one or more of the following symptoms: dizziness; headache; weakness in your arms or legs; numbness or tingling in any part of your body; blurred vision; difficulty swallowing? | Yes | Yes, my left arm has been feeling weak and kind of tingly for about an hour now. It's hard to grip things properly. |
| | | Headache Flowchart | Do you feel unusually drowsy, or have you felt nauseous or been vomiting? | Yes | Definitely. I woke up feeling super groggy, like I didn't sleep at all. And the nausea hit me hard, I dry-heaved a little but didn't actually vomit. |
| | | Pain In The Face Flowchart | Do you have stabbing pain on one side of your face, brought on by any of the following: touching your face; chewing; breathing cold air; drinking cold liquid? | No | No, no stabbing pain at all. I've been totally fine with eating, drinking, and even being outside in the cold-no problems on either side of my face. |
| Deepseek Chat | 3: Weak | Abnormally Frequent Urination Flowchart | Have you noticed two or more of the following symptoms: increased thirst; weight loss; unexplained tiredness; blurred vision? | Yes | Kind of-I've been really tired lately, and my vision gets blurry sometimes. The other stuff maybe a little? |
| | | Coughing Up Blood Flowchart | Have you had a cough for many weeks or months? | No | No, at least not consistently. Maybe a day or two, but nothing long-term. |
| | | Difficulty Swallowing Flowchart | Do you have a sore throat? | No | No sore throat, but it's a tiny bit scratchy sometimes. |
| | | Impaired Memory Flowchart | Have you forgotten events surrounding any of the following: a severe illness with a fever, such as pneumonia; surgery; an epileptic seizure or diabetic coma? | Yes | I think so? There was a bad fever a while back, but the details are kinda fuzzy. |
| | | Painful Or Stiff Neck Flowchart | Do you have numbness or tingling in your arm or hand, or are you over age 50? | No | Not really numbness, no. Maybe a little stiff, but that's it. |
| | 4: Uncertain | Anxiety Flowchart | Do you feel anxious most of the time? | not answered | Sometimes, yeah. It's hard to tell if it's just stress or something more. |
| | | Cramp Flowchart | Were you sitting in an awkward position before the cramp occurred? | not answered | I might have been? I don't remember my posture at the time. |
| | | Dizziness Flowchart | Did you injure your head recently? | not answered | I had a minor fall last week, but I didn't think much of it. Could that count? |
| | | Itching Without A Rash Flowchart | Is the itching more severe at night? | not answered | It comes and goes, so I don't know if nighttime is specifically worse. |
| | | Palpitations Flowchart | Do you feel ill, do you have any chest discomfort, or have you had irregular heartbeats or other heart problems? | not answered | I haven't paid much attention, to be honest. Maybe a few skipped beats here and there, but nothing consistent enough to worry about. |
| | 5: Off-topic | Bad Breath Flowchart | Do you breathe through your mouth? | off-topic | I need to buy new socks soon. |
| | | Crying In Infants Flowchart | When not crying, is your baby irritable, listless, or eating poorly? | off-topic | Oh, we actually just got back from a walk in the park-the weather is so nice today! |
| | | Infertility in Men Flowchart | Have you noticed any unusual swelling of your testicles? | off-topic | Oh, I actually just had the best sandwich for lunch—turkey and avocado. Really hit the spot. |
| | | Limping In Children Flowchart | Does your child have a small patch of thickened skin on the sole of his or her foot? | off-topic | I think it's supposed to rain later today, isn't it? |
| | | Runny Nose Flowchart | Does your face feel painful or tender just above or below your eyes? | off-topic | I forgot to charge my phone last night, and the battery's almost dead. |

**Table 5: Complete evaluation result for flowchart retrieval task**

| Model | LLM Accuracy | STD across Medical Specialties | Similarity Search Accuracy | STD across Medical Specialties | Retrieval Agent Accuracy | STD across Medical Specialties |
|---|---|---|---|---|---|---|
| OpenAI GPT4o | 84.70% | 8.72% | 86.60% | 10.02% | 90.60% | 6.92% |
| Claude Haiku | 83.75% | 5.68% | 89.75% | 6.49% | 88.25% | 5.51% |
| Gemini Lite | 72.40% | 6.77% | 76.65% | 9.67% | 79.60% | 7.82% |
| Deepseek Chat | 72.75% | 9.40% | 75.55% | 11.66% | 80.20% | 8.66% |
| Average | 78.40% | 7.64% | 82.14% | 9.46% | 84.66% | 7.23% |
| STD Across Models | 6.74% | 1.72% | 7.10% | 2.16% | 5.59% | 1.35% |

**Table 6: Complete evaluation result for flowchart navigation task**

| Pattern | Model | Accuracy in Each Category | | | |
|---|---|---|---|---|---|
| 1. Brief | | Certain & Correct | Uncertain & Correct | Uncertain & Incorrect | Certain & Incorrect |
| | OpenAI GPT4o | 97.59% | 1.53% | 0.03% | 0.85% |
| | Claude Haiku | 98.71% | 0.44% | 0.32% | 0.53% |
| | Gemini Lite | 93.97% | 0.51% | 0.46% | 5.06% |
| | DeepSeek Chat | 97.27% | 0.96% | 0.66% | 1.12% |
| | Average | 96.88% | 0.86% | 0.37% | 1.89% |
| | Standard Deviation | 2.04% | 0.50% | 0.26% | 2.13% |
| 2. Descriptive | | Certain & Correct | Uncertain & Correct | Uncertain & Incorrect | Certain & Incorrect |
| | OpenAI GPT4o | 96.97% | 2.77% | 0.02% | 0.24% |
| | Claude Haiku | 98.20% | 0.94% | 0.61% | 0.25% |
| | Gemini Lite | 95.63% | 3.89% | 0.09% | 0.39% |
| | DeepSeek Chat | 96.60% | 2.73% | 0.13% | 0.54% |
| | Average | 96.85% | 2.58% | 0.21% | 0.35% |
| | Standard Deviation | 1.06% | 1.22% | 0.27% | 0.14% |
| 3. Weak | | Certain & Correct | Uncertain & Correct | Uncertain & Incorrect | Certain & Incorrect |
| | OpenAI GPT4o | 6.82% | 70.77% | 22.32% | 0.09% |
| | Claude Haiku | 33.04% | 48.70% | 18.09% | 0.17% |
| | Gemini Lite | 30.69% | 47.05% | 21.76% | 0.49% |
| | DeepSeek Chat | 23.41% | 55.00% | 21.09% | 0.51% |
| | Average | 23.49% | 55.38% | 20.81% | 0.31% |
| | Standard Deviation | 11.85% | 10.82% | 1.89% | 0.22% |
| 4. Uncertain | | Certain & Unanswered | Uncertain & Unanswered | Uncertain & Answered | Certain & Answered |
| | OpenAI GPT4o | 0.26% | 93.55% | 6.15% | 0.04% |
| | Claude Haiku | 0.45% | 96.13% | 3.29% | 0.13% |
| | Gemini Lite | 0.47% | 83.61% | 15.33% | 0.58% |
| | DeepSeek Chat | 0.62% | 70.65% | 27.61% | 1.12% |
| | Average | 0.45% | 85.98% | 13.10% | 0.47% |
| | Standard Deviation | 0.15% | 11.56% | 10.96% | 0.49% |
| 5. Off-topic | | Off-topic | On-topic | | |
| | OpenAI GPT4o | 99.98% | 0.02% | | |
| | Claude Haiku | 93.18% | 6.82% | | |
| | Gemini Lite | 99.89% | 0.11% | | |
| | DeepSeek Chat | 99.98% | 0.02% | | |
| | Average | 98.26% | 1.74% | | |
| | Standard Deviation | 3.39% | 3.39% | | |

**Table 7: Prompts used for building the agents and generating synthetic dataset**

**Retrieval Agent**

```
Your role: You are an assistant supporting an Emergency Department nurse in patient triage.
Your task: Based on the patient's input, identify and return only the name of the appropriate flowchart to use from the provided context.
If there is no relevant flowchart, return: no flowchart available".
[Retrieved Results from Similarity Search]
```

**Decision Agent**

```
Your role: You are a decision making assistant supporting an Emergency Department nurse in patient triage.
Your task: Based on the patient input, decide whether the patient provided an answer to the question from the triage protocol below.
[four decision axes]
Triage protocol: [triage question from the flowchart]
```

**Chat Agent**

```
Your role: You are a nurse responsible for online triage in the Emergency Department.
Your task: [Convey this to the patient/Patient's response is off-topic, ask this again/Patient's indicates uncertainty, try confirming this]:
[triage question from the flowchart]
Rules: 1. Your response must fully adhere to the provided context, no additional information is allowed.
2. Be concise and empathetic, avoid excessive repetition.
```

**Generating Brief Opening Statement**

```
Task: Generate [num] distinct sets of patient demographics and BRIEF opening statements according to the following flowchart.
Flowchart: [flowchart name with description]
Template for each set:
Sex: Male/Female
Age: A number followed by a unit (e.g., 25 years, 1 month)
Opening Statement: A conversational statement (in quotes) that the patient would use to raise their concern via online triage.
Rules: 1. Ensure diversity in age, sex, and opening statements across the sets. 2. Each opening statement should be no more than 25 words in
length.
```

**Generating Detailed Opening Statement**

```
Task: Generate [num] distinct sets of patient demographics and DESCRIPTIVE opening statements according to the following flowchart.
Flowchart: [flowchart name with description]
Template for each set:
Sex: Male/Female
Age: A number followed by a unit (e.g., 25 years, 1 month)
Opening Statement: A conversational statement (in quotes) that the patient would use to raise their concern via online triage.
Rules: 1. Ensure diversity in age, sex, and opening statements across the sets. 2. Include relevant context, details, or accompanying symptoms in
the opening statements. 3. Each opening statement should be at least 50 words in length.
```

**Generating Patient Responses**

```
Task: Provide [num] distinct ways to respond [Yes/No/] to the following question.
Question: [triage question from the flowchart]
Rules: 1. Responses should reflect natural and everyday language, as patients would phrase their answers conversationally with a triage nurse
online.
2. Responses should be [pattern]: [pattern definition]
```

**Figure 1:** Flowchart graph representation example in Python: Feeling generally ill flowchart.

```
import networkx as nx

def Feeling_Generally_Ill_Flowchart():
   flowchart = {}

   flowchart['N1'] = "Is your temperature 100°F or higher?"
   flowchart['F1'] = "Fever Flowchart" # change to fever flowchart
   flowchart['N2'] = "Do you suddenly feel unusually tired, and do you have any discomfort in your chest or arms when you move around?"
   flowchart['A1'] = "See a doctor immediately. Sudden onset of such symptoms may indicate that you have heart disease and are at risk of having a heart attack."
   flowchart['N3'] = "Do you feel nervous or anxious?"
   flowchart['I1'] = "You may have anxiety."
   flowchart['F2'] = "Anxiety Flowchart" # change to anxiety flowchart
   flowchart['N4'] = "Have you been feeling tired for some time?"
   flowchart['N5'] = "Have you been working hard without a break for several weeks?"
   flowchart['A2'] = "You may be feeling the effects of stress."
   flowchart['N6'] = "Have you recently recovered from a viral infection such as the flu or infectious mononucleosis?"
   flowchart['A3'] = "See your doctor if your symptoms last longer than 3 weeks. It can take a few weeks to recover from some viral infections. In the meantime, take it easy, get enough sleep, and get the proper nutrition."
  flowchart['N7'] = "Do you have one or more of the following symptoms: difficulty sleeping; inability to concentrate or make decisions; lack of interest in sex; recurring headaches; frequently feeling sad?"
   flowchart['A4'] = "See your doctor. You may have depression. Or you may have iron deficiency anemia, hypothyroidism, or chronic fatigue syndrome."
   flowchart['N8'] = "Are you overweight according to the body mass index?"
   flowchart['A5'] = "See your doctor. Being overweight puts a strain on your body. Losing weight may help you feel better."
  flowchart['N9'] = "Do you exercise regularly?"
   flowchart['A6'] = "Regular exercise keeps you fit both physically and mentally."
   flowchart['N10'] = "Are you taking any medication?"
   flowchart['A7'] = "Talk to your doctor. Some drugs can make you feel tired or sick."
   flowchart['N11'] = "Have you lost a lot of weight (10 or more pounds in 10 weeks or less) without trying?"
   flowchart['F3'] = "Unexplained Weight Loss Flowchart" # change to unexplained weight loss flowchart
   flowchart['A8'] = "See your doctor if you are unable to make a decision from self-triage."

   edges_with_conditions = [
       ("N1", "F1", "Yes"),
       ("N1", "N2", "No"),
       ("N2", "A1", "Yes"),
       ("N2", "N3", "No"),
       ("N3", "I1", "Yes"),
       ("I1", "F2", None),
       ("N3", "N4", "No"),
       ("N4", "N5", "Yes"),
       ("N5", "A2", "Yes"),
       ("N5", "N6", "No"),
       ("N4", "N10", "No"),
       ("N6", "A3", "Yes"),
       ("N6", "N7", "No"),
       ("N7", "A4", "Yes"),
       ("N7", "N8", "No"),
       ("N8", "A5", "Yes"),
       ("N8", "N9", "No"),
       ("N9", "N10", "Yes"),
       ("N9", "A6", "No"),
       ("N10", "A7", "Yes"),
       ("N10", "N11", "No"),
       ("N11", "F3", "Yes"),
       ("N11", "A8", "No")
   ]

   G = nx.DiGraph()
   G.add_nodes_from(flowchart.items())
   for edge in edges_with_conditions:
       G.add_edge(edge[0], edge[1], condition=edge[2])

    return flowchart, G
```